\documentclass[mnsc,nonblindrev]{informs3_hide}
\OneAndAHalfSpacedXI 

\usepackage[english]{babel}
\usepackage[autostyle, english = american]{csquotes}
\MakeOuterQuote{"}
\usepackage[colorlinks,linkcolor=blue, citecolor=blue]{hyperref}
\usepackage[normalem]{ulem}

\usepackage{cleveref}
\crefname{subsection}{section}{subsections}
\usepackage{eqnarray}
\usepackage[ruled,vlined,linesnumbered]{algorithm2e}
\usepackage{amsmath, bbm, enumitem, xparse, bm, lipsum}
\usepackage{amssymb}

\Crefname{algocf}{Algorithm}{Algorithms}

\usepackage{graphicx}
\usepackage{subfigure, epsfig}
\usepackage{natbib}
\usepackage{appendix}
\usepackage{url}

\usepackage{xfrac}
\usepackage{xcolor}
\usepackage{bbm}
\usepackage{booktabs}
\usepackage{tikz}
\usepackage{subcaption}
\usepackage{float}
\newtheorem{lemma}{Lemma}

\newtheorem{theorem}{Theorem}
\newtheorem{example}{Example}
\newenvironment{myprocedure}[1][htb]
{%
  \begin{algorithm}[#1]%
}{%
  \end{algorithm}%
}

\DeclareRobustCommand{\mybox}[2][gray!10]{%
\begin{tcolorbox}[   
        left=0pt,
        right=0pt,
        top=0pt,
        bottom=0pt,
        colback=#1,
        colframe=#1,
        enlarge left by=0mm,
        boxsep=15pt,
        arc=5pt,outer arc=5pt,
        ]
        #2
\end{tcolorbox}
}

\DeclareRobustCommand{\myboxtwo}[2][gray!15]{
\begin{tcolorbox}[ 
        colback=white,      
        colframe=gray,  
        boxrule=0.2pt,      
        arc=2pt,outer arc=2pt,
        left=12pt,
        right=12pt,
        top=5pt,
        bottom=5pt,
        width=1.07\linewidth,
        enlarge left by=-0.55cm,
        before upper=\renewcommand{\baselinestretch}{1.3}\selectfont,
        after upper=\normalfont
        ]
 #2
 \end{tcolorbox}
}

\usepackage{thm-restate}

\usepackage[breakable, skins]{tcolorbox}
\DeclareRobustCommand{\mybox}[2][gray!10]{%
\begin{tcolorbox}[
        left=0.5pt,
        right=0.5pt,
        top=0.5pt,
        bottom=0.5pt,
        colback=#1,
        colframe=#1,
        width=\dimexpr\textwidth\relax, 
        enlarge left by=1mm,
        boxsep=3pt,
        arc=2pt,outer arc=2pt,
        ]
        #2
        
\end{tcolorbox}
}
\usepackage{csquotes}
\makeatletter
\renewenvironment*{displayquote}
  {\begingroup\setlength{\leftmargini}{0.2cm}\csq@getcargs{\csq@bdquote{}{}}}
  {\csq@edquote\endgroup}
\makeatother

\newtheorem{assumption}{Assumption}

\usepackage{mathtools, bm, amsmath}

\definecolor{cornellred}{rgb}{0.7, 0.11, 0.11}
\definecolor{maroon}{rgb}{0.52, 0, 0}
\definecolor{dgreen}{rgb}{0.0, 0.5, 0.0}
\definecolor{ballblue}{rgb}{0.13, 0.67, 0.8}
\definecolor{royalbamsmathlue(web)}{rgb}{0.25, 0.41, 0.88}
\definecolor{bleudefrance}{rgb}{0.19, 0.55, 0.91}
\definecolor{royalazure}{rgb}{0.0, 0.22, 0.66}
\usepackage{hyperref}
\hypersetup{
	colorlinks = true,
	linkcolor=royalazure,
	urlcolor=royalazure,
	citecolor=royalazure,
	linkbordercolor = {white}
}
\usepackage{cleveref}
\usepackage{enumitem}
\usepackage{multirow}
\Crefname{assumption}{Assumption}{Assumptions}

\usepackage{pgf,tikz,pgfplots}
\pgfplotsset{compat=1.15}
\usetikzlibrary{arrows}
\usepackage{pgfplots}
\usetikzlibrary{patterns}
\usepgfplotslibrary{fillbetween}
\usetikzlibrary{intersections}

\usepackage{fix-cm}

\usetikzlibrary{arrows, decorations.markings}

\tikzstyle{vecArrow} = [thick, decoration={markings,mark=at position
	1 with {\arrow[semithick]{open triangle 60}}},
	double distance=1.4pt, shorten >= 5.5pt,
	preaction = {decorate},
postaction = {draw,line width=1.4pt, white,shorten >= 4.5pt}]
\tikzstyle{innerWhite} = [semithick, white,line width=1.4pt, shorten >= 4.5pt]

	\newcommand{\reals}{\mathbb{R}}

\makeatletter
\newcommand{\subalign}[1]{%
  \vcenter{%
    \Let@ \restore@math@cr \default@tag
    \baselineskip\fontdimen10 \scriptfont\tw@
    \advance\baselineskip\fontdimen12 \scriptfont\tw@
    \lineskip\thr@@\fontdimen8 \scriptfont\thr@@
    \lineskiplimit\lineskip
    \ialign{\hfil$\m@th\scriptstyle##$&$\m@th\scriptstyle{}##$\hfil\crcr
      #1\crcr
    }%
  }%
}
\makeatother

	\providecommand{\given}{}
	\DeclarePairedDelimiterX{\set}[1]\{\}{\renewcommand\given{\nonscript\:\delimsize\vert\nonscript\:\mathopen{}}#1}
	\let\Pr\relax
	\DeclarePairedDelimiterXPP{\Pr}[1]{\mathbb{P}}[]{}{\renewcommand\given{\nonscript\:\delimsize\vert\nonscript\:\mathopen{}}#1}
	\DeclarePairedDelimiterXPP{\Ex}[1]{\mathbb{E}}[]{}{\renewcommand\given{\nonscript\:\delimsize\vert\nonscript\:\mathopen{}}#1}

\newcommand{\xhdr}[1]{\vspace{2mm} \noindent{\bf #1}}
 
\newcommand{\xbf}{\mathbf{x}}

\newcommand*{\rom}[1]{\expandafter\romannumeral #1}
\newcommand{\Rom}[1]{\uppercase\expandafter{\romannumeral #1\relax}}

\newcommand{\reward}{r}
\newcommand{\sreward}{\hat{r}}

\newcommand{\numResource}{m}
\newcommand{\numPeriod}{T}
\newcommand{\numSize}{n}
\newcommand{\cdf}{F}
\newcommand{\pdf}{f}
\newcommand{\cdfInverse}{F^{-1}}
\newcommand{\empcdf}{\bar{F}}
\newcommand{\estcdf}{\hat{F}}
\newcommand{\estcdfInverse}{\hat{F}^{-1}}

\newcommand{\typedf}{P}

\newcommand{\query}{t}
\newcommand{\type}{j}
\newcommand{\stype}{\hat{j}}
\newcommand{\instance}{I}
\newcommand{\sizeset}{\mathcal{A}}
\newcommand{\lowreward}{\underline{r}}
\newcommand{\uppreward}{\bar{r}}

\newcommand{\offDecision}{x^{\text{off}}}
\newcommand{\piDecision}{x^{\pi}}
\newcommand{\regret}{\text{Regret}}
\newcommand{\snumQuery}{\hat{d}}
\newcommand{\numQuery}{d}
\newcommand{\tnumQuery}{b}
\newcommand{\tsnumQuery}{\hat{b}}
\newcommand{\sample}{N}
\newcommand{\bnumQuery}{\bm{d}}
\newcommand{\bsnumQuery}{\hat{\bm{d}}}
\newcommand{\bconstraint}{\bm{C}}
\newcommand{\constraint}{C}
\newcommand{\brconstraint}{\bm{c}}
\newcommand{\rconstraint}{c}
\newcommand{\consumption}{\text{cons}}
\newcommand{\myopic}{\text{Myopic}}

\newcommand{\tV}{\tilde{V}}
\newcommand{\bV}{\bar{V}}
\newcommand{\hV}{\Hat{V}}

\newcommand{\Voff}{V^{\mathrm{off}}}

\newcommand{\bVsemi}{\bar{V}^{\mathrm{semi}}}
\newcommand{\Vsemi}{V^{\mathrm{semi}}}

\newcommand{\bVfluid}{\bar{V}^{\mathrm{fld}}}
\newcommand{\Vfluid}{V^{\mathrm{fld}}}
\newcommand{\hL}{\Hat{L}}

\newcommand{\hG}{\Hat{G}}
\newcommand{\hq}{\Hat{q}}

\newcommand{\bq}{\bm{q}}
\newcommand{\blambda}{\bm{\lambda}}
\newcommand{\hlambda}{\Hat{\lambda}}
\newcommand{\tlambda}{\tilde{\lambda}}
\newcommand{\tblambda}{\tilde{\bm{\lambda}}}
\newcommand{\hblambda}{\Hat{\bm{\lambda}}}
\newcommand{\amax}{a_{\text{max}}}

\newcommand{\Var}{\mathrm{Var}}

\newcommand{\size}{\bm{a}}

\newcommand{\ba}{\bm{a}}

\newcommand{\tq}{\tilde{q}}

\newcommand{\FATP}{\textsc{Fully Adaptive Threshold Policy}}
\newcommand{\STP}{\textsc{Static Threshold Policy}}
\newcommand{\PATP}{\textsc{Partial Adaptive Threshold Policy}}

\newcommand{\prob}[2][]{\text{Pr}\ifthenelse{\not\equal{}{#1}}{_{#1}}{}\!\left[{\def\givenn{\middle|}#2}\right]}
\newcommand{\expect}[2][]{\mathbb{E}\ifthenelse{\not\equal{}{#1}}{_{#1}}{}\!\left[{\def\givenn{\middle|}#2}\right]}

\newcommand{\tparen}{\big}
\newcommand{\tprob}[2][]{\text{Pr}\ifthenelse{\not\equal{}{#1}}{_{#1}}{}\tparen[{\def\given{\tparen|}#2}\tparen]}
\newcommand{\texpect}[2][]{\mathbb{E}\ifthenelse{\not\equal{}{#1}}{_{#1}}{}\tparen[{\def\given{\tparen|}#2}\tparen]}

\newcommand{\sprob}[2][]{\text{Pr}\ifthenelse{\not\equal{}{#1}}{_{#1}}{}[#2]}
\newcommand{\sexpect}[2][]{\mathbb{E}\ifthenelse{\not\equal{}{#1}}{_{#1}}{}[#2]}

\newcommand{\dd}{{\mathrm d}}

\newcommand{\indicator}[1]{{\mathbbm{1}\left\{ #1 \right\}}}

\usepackage[dvipsnames]{xcolor}

\usepackage{xparse}

\usepackage{natbib,bbm,multirow,multicol}
 \bibpunct[, ]{(}{)}{,}{a}{}{,}%
 \def\bibfont{\small}%

\author{
 Yiding Feng \\
  Department of Industrial Engineering and Decision Analytics\\
  Hong Kong University of Science and Technology\\
  \texttt{ydfeng@ust.hk} \\
   \And
 Jiashuo Jiang \\
  Department of Industrial Engineering and Decision Analytics\\
  Hong Kong University of Science and Technology\\
  \texttt{jsjiang@ust.hk} \\
  \And
 Yige Wang \\
  Department of Industrial Engineering and Decision Analytics\\
  Hong Kong University of Science and Technology\\
  \texttt{ywangrc@connect.ust.hk} \\
}

\begin{document}

\RUNAUTHOR{Feng, Jiang, Wang}

\RUNTITLE{Non-Stationary Online Resource Allocation: Learning from a Single Sample}

\TITLE{
Non-Stationary Online Resource Allocation: Learning from a Single Sample
}

\ARTICLEAUTHORS{%
\AUTHOR{$\text{Yiding Feng}^{\dag}$, $\text{Jiashuo Jiang}^{\dag}$, $\text{Yige Wang}^{\dag}$}

\AFF{\  \\
$\dag~$Department of Industrial Engineering \& Decision Analytics, Hong Kong University of Science and Technology
}
}

\ABSTRACT{
We study online resource allocation under non-stationary demand with a minimum offline data requirement. In this problem, a decision-maker must allocate multiple types of resources to sequentially arriving queries over a finite horizon. Each query belongs to a finite set of types with fixed resource consumption and a stochastic reward drawn from an unknown, type-specific distribution. Critically, the environment exhibits arbitrary non-stationarity---arrival distributions may shift unpredictably---while the algorithm requires only \emph{one historical sample per period} to operate effectively. We distinguish two settings based on sample informativeness: (i) \emph{reward-observed samples} containing both query type and reward realization, and (ii) the more challenging \emph{type-only samples} revealing only query type information.

We propose a novel type-dependent quantile-based meta-policy that decouples the problem into modular components: reward distribution estimation, optimization of target service probabilities via fluid relaxation, and real-time decisions through dynamic acceptance thresholds. For reward-observed samples, our static threshold policy achieves $\tilde{O}(\sqrt{T})$ regret without requiring large-budget assumptions. For type-only samples, we first establish that sublinear regret is impossible without additional structure; under a mild minimum-arrival-probability assumption, we design both a partially adaptive policy attaining the same $\tilde{O}({T})$ bound and, more significantly, a fully adaptive resolving policy with careful rounding that achieves the first poly-logarithmic regret guarantee of $O((\log T)^3)$ for non-stationary multi-resource allocation. Our framework advances prior work by operating with minimal offline data (one sample per period), handling arbitrary non-stationarity without variation-budget assumptions, and supporting multiple resource constraints---demonstrating that near-optimal performance is attainable even under stringent data requirements.
}
\maketitle


\section{Introduction}
\label{sec:intro}
Online resource allocation under uncertainty is a fundamental problem in sequential decision-making, with broad applications in revenue management, digital advertising, cloud infrastructure, and sharing economy platforms. In this paradigm, a decision-maker must allocate limited resources to sequentially arriving queries over a finite time horizon. Each query yields a stochastic reward and consumes a random vector of resources. Decisions must be made immediately and irrevocably upon arrival without knowledge of future queries, with the objective of maximizing total reward subject to resource constraints.

Traditional models for online resource allocation often rely on idealized assumptions, which commonly presuppose \emph{stationary demand processes}. In practice, however, operational environments increasingly defy these conditions.
Consider an e-commerce cloud platform anticipating a flash sale: demand may surge unpredictably due to viral social trends or seasonal peaks such as Singles’ Day. Similarly, a ride-hailing service entering a new city faces rapidly shifting trip patterns influenced by weather, local events, or evolving commuter habits. In digital advertising, a breaking news event can abruptly reshape user engagement, rendering pre-event click-through models obsolete. 
These scenarios collectively underscore a fundamental tension in modern resource allocation: \emph{environments exhibit dynamic non-stationarity}, characterized by trends, seasonality, and external shocks. Consequently, classical algorithms designed for stationary settings often exhibit degraded reliability and limited adaptability when deployed in highly volatile conditions.

Given the inherent intractability of computing optimal policies under uncertainty, the field prioritizes algorithms with rigorous performance guarantees. The standard benchmark is \emph{regret}—the expected difference between the cumulative reward of online policy and that of optimal offline policy with full knowledge of all future realizations beforehand. Although research in stationary settings has achieved near-optimal theoretical results, prior work has also shown that in online learning problems, sublinear regret is unattainable without a minimal amount of data. Therefore, bridging the gap between idealized theoretical assumptions and realities of non-stationarity and data scarcity forms the core motivation for our main research question:
\myboxtwo{
\begin{displayquote}
\emph{How can we design online allocation algorithms that learn effectively from sparse, non-stationary data to achieve near-optimal performance—ideally with sublinear, or even logarithmic regret?}
\end{displayquote}
}

In this work, we investigate non‑stationary online resource allocation with limited historical samples. We consider a finite‑horizon setting with multiple resource constraints, where queries arrive sequentially. Each query belongs to a finite set of types, and every type is associated with a fixed resource consumption vector and a stochastic reward drawn from an unknown, continuous, type-specific distribution. In each period, the decision-maker observes a query's type and its random reward realization, then must decide irrevocably whether to accept (consuming resources for a reward) or reject it. The arrival process is non-stationary, with distributions that can change arbitrarily over time. Critically, both the arrival distribution of types and the conditional reward distributions are initially unknown; the algorithm has access to only one independent historical sample per period in advance, which may lack reward information. The objective is to design an online policy that leverages these sparse samples to adapt to distributional shifts while maximizing cumulative reward subject to resource constraints.

\subsection{Our Main Results and Contributions}
In this paper, we propose a novel, unified framework that achieves strong theoretical guarantees under non-stationarity and extreme data scarcity. A cornerstone of our contribution is the design of near-optimal online allocation algorithms under the minimal possible data requirement: leveraging only a single historical sample per period to adapt to arbitrary distributional shifts while attaining sublinear or even logarithmic regret. Our main contributions are as follows:

\xhdr{Theoretical Regret Guarantees:} 
For the online non-stationary multi-resource allocation problems with unknown type-arrival and reward distributions, we design an fully adaptive algorithm that achieves poly-logarithmic regret using only a single historical sample per period—containing solely type-arrival information. Our main result is as follows:
\mybox{\emph{\textbf{Main Result:} We establish the first poly-logarithmic regret upper bound $O((\log\numPeriod)^3)$ for online multi-resource allocation problems under non-stationarity with only one type-only historical sample per period.}}

To the best of our knowledge, this result provides the first poly-logarithmic regret guarantee for non-stationary online resource allocation. Our bound relies on a minimal-arrival-probability assumption. We further demonstrate the necessity of this assumption by constructing a counterexample, which shows that without such assumption, sublinear regret is unattainable in the worst case when only type samples are available. Under the same conditions, we also derive a $\tilde{O}(\sqrt{\numPeriod})$ regret bound using a partially adaptive algorithm that updates reward distribution estimates online that is easier to implement. If the minimal-arrival-probability assumption is relaxed, attaining sublinear regret requires access to historical reward information. In this setting, we propose a more easier static threshold policy with reward-observed samples that achieves the same $\tilde{O}(\sqrt{\numPeriod})$ upper bound.

The closest work to ours is by \citet{GSW-25}, who also studied multi-resource allocation with a single historical sample and proposed an exponential pricing algorithm that attains a $(1-\epsilon)$ approximation to the hindsight optimum. Their result considers a large-budget assumption-specifically, budgets of order $\tilde{\Omega}(1/\epsilon^6)$-which, when resource scales linearly with time horizon, translates to a regret rate of $\tilde{O}(\numPeriod^{5/6})$.
In contrast, our quantile-based method operates under arbitrary resource levels and achieves a sharper $\tilde{O}(\sqrt{\numPeriod})$ regret under the same setting with reward-observed samples. For the simpler single-resource setting, \citet{BKMSW-23} obtains a $\tilde{O}(\sqrt{\numPeriod})$ bound, which our work matches in the multi-resource case under milder assumptions.

Overall, our approach relaxes several critical assumptions in prior work and delivers rigorous, scalable performance guarantees in data‑scarce and non‑stationary environments. We advance the theoretical understanding of online resource allocation through a novel and flexible quantile-based framework that unifies the analysis across different distributional assumptions, as will be elaborated later.

\xhdr{A Meta Quantile-Based Algorithm Design:} 
To achieve the theoretical guarantees described, we introduce a novel \emph{type‑dependent, quantile‑based policy framework}. This design departs fundamentally from the conventional dual‑based paradigm by modularly decomposing the online allocation problem into three distinct components: (i) Reward Function Estimation: a standalone learning sub‑problem where state‑of‑the-art methods can be applied directly; (ii) Optimization of Target Service Probabilities: a strategic layer that resolves the long-term resource-reward trade-off; and (iii) Real‑time Decision‑making: executed via dynamic type-specific acceptance thresholds derived from estimated reward quantiles.

The power of this decomposition lies in two key advantages:
(i) Modularity for Direct Technical Integration: It establishes a clean interface between high-level resource budgeting and per-period operations. The framework is explicitly designed to be modular, allowing any advancements in quantile or distribution estimation to be plugged directly into the policy without modifying the core allocation logic.
(ii) Type-Dependent, Non-Interfering Decisions: Each query type is managed independently through its own quantile trajectory and acceptance threshold. This type-dependent design ensures that the decision logic for one type does not interfere with or complicate the processing of another, leading to a more transparent and robust management of the resource-reward trade-off across diverse arrival types.

In contrast to prior dual-based approaches—which typically adopt specialized, monolithic designs that tightly couple dual variable learning with allocation logic and rely on intricate, algorithm-specific analyses—our quantile-based framework cleanly decouples learning from optimization. For instance, \citet{BKMSW-23} employ dual variables as adaptive shadow prices updated via Dual FTRL algorithm to control budget pacing under uncertainty, while \citet{GSW-25} dynamically adjust dual prices using an exponential rule to guide resource allocation. Although effective, such methods inherently intertwine dual-space learning with decision-making, complicating adaptation and analysis.
Our approach operates fundamentally differently: it bypasses explicit dual variable maintenance altogether. Allocation decisions are derived solely by solving the primal optimization problem using locally estimated reward quantiles specific to each query type. This design requires only minimal, interpretable information per type, eliminates dependencies on dual dynamics, and enhances both theoretical tractability and practical adaptability across diverse online allocation settings—offering a versatile alternative to tightly coupled dual-based paradigms.

We position our contributions against key related works in the following \Cref{tab:comparison} under our formulation where some mild assumptions are omitted for simplicity.

\begin{table}[ht]
\centering
\caption{Comparison of our work to key related literature}
\label{tab:comparison}
\renewcommand{\arraystretch}{1.3}
\begin{tabular}{cccc}
\toprule
\textbf{} & \textbf{resource number} & \textbf{method} & \textbf{regret} \\
\midrule
\multirow{2}{*}{\cite{BKMSW-23}} & \multirow{2}{*}{single} & \multirow{2}{*}{dual-based} & \multirow{2}{*}{$\tilde{O}(T^{1/2})$} \\
& & & \\
\midrule
\multirow{2}{*}{\cite{GSW-25}} & \multirow{2}{*}{multiple} & \multirow{2}{*}{dual-based} & \multirow{2}{*}{$\tilde{O}(T^{5/6})$} \\
& & & \\
\midrule
\multirow{2}{*}{Our work} & \multirow{2}{*}{multiple} & \multirow{2}{*}{quantile-based} & \multirow{2}{*}{$O((\log T)^3)$, $\tilde{O}(T^{1/2})$} \\
& & & \\
\bottomrule
\end{tabular}
\end{table}

\subsection{Other Related Literature}
\xhdr{Network Revenue Management with Logarithmic Regret:}
Network Revenue Management (NRM) is a central and extensively studied problem in online resource allocation, where a key objective is to develop practically effective policies with strong theoretical guarantees. 
The field's foundations include the dynamic pricing model for NRM introduced by \citet{GV-97}. A seminal contribution by \citet{TV-98} proposed a static bid-price policy based on the dual variables of an ex-ante fluid relaxation, establishing a sublinear regret bound. This approach was later refined by \citet{RM-08}, who demonstrated that periodically re-solving the fluid program to update bid-prices yields an improved regret bound of $o(\sqrt{T})$.
When the underlying demand functions are unknown, the problem requires balancing learning with optimization. \citet{BZ-12} designed "blind" pricing policies that explore and exploit based only on observed sales data. Subsequent work, such as that of \citep{FSW-18}, employed Bayesian methods like Thompson sampling to address this trade-off under inventory constraints. In a different direction, \citet{DJSW-19} developed a single algorithm that attains a $(1-\epsilon)$ fraction of the offline optimum for every possible arrival distribution. Recent extensions have addressed more complex settings, such as reusable resources \citep{BM-22} and non-stationary environments with imperfect distributional knowledge \citep{JLZ-25}.

A distinct and theoretically significant line of research aims to establish tighter, often logarithmic, regret bounds. These results typically require more structured problem assumptions, such as discrete distributions with finite support or non-degeneracy assumptions. For instance, \citet{JK-12} analyzed certainty-equivalent heuristics for NRM with customer choice, providing bounded revenue loss. \citet{BW-20} designed a re-solving heuristic with a constant regret bound independent of the time horizon and resource capacities. More recently, \citet{LY-22} derived logarithmic regret for online linear programming under local strong convexity, and a result was further tightened and extended by \citet{Bra-25}. While \citet{JMZ-25} introduced a similar quantile-based policy that get rid of the dependence of non-degeneracy, our framework significantly extends the results by accommodating arbitrary non‑stationary arrival processes and incorporating online distribution learning for settings with initially unknown rewards.

\xhdr{Online Learning with Samples:}
The field of online learning with prior samples investigates how access to historical data can enhance sequential decision-making. Early foundational work often assumed inputs were drawn from a known or partially known distribution. For example, \citet{GGLS-08} analyzed online algorithms under inputs from a fixed distribution, while \citet{HZ-09} considered sequences from non-identical distributions. Other studies designed robust algorithms to mitigate overfitting in problems such as the knapsack secretary problem \citep{BGSZ-19}, or developed sample-driven methods for optimal stopping under random-order arrivals \citep{CCES-24}.
More recent research has shifted toward stringent data limitations, focusing on online learning with very few historical sample. This line of inquiry has been applied across various online decision problems: prophet inequalities have been studied under single-sample access in \citep{AKW-14,CDFFLLPPR-22,CZ-24}; online matching was examined by \citet{KNR-22}; and online network revenue management was considered by \citet{AFGS-22}. Furthermore, \citet{DKLRS-24} investigated online combinatorial allocation with few samples for bidders with combinatorial valuations. Collectively, this work demonstrates both the challenge and the feasibility of achieving near-optimal performance with minimal prior data.

Several studies are particularly pertinent to our setting. \citet{BSX-20} studied online pricing with offline data, but their approach relies on $n$ samples per product and assumes linear demand models. \citet{CL-25} considered an episodic framework repeated over $H$ rounds, which uses multiple samples per episode and is confined to single-resource settings. In contrast, our framework requires only one sample per period and accommodates general, nonparametric reward structures with multiple resource setting.
This context underscores our core contribution: we develop a framework for multi-resource online allocation that operates under extreme data scarcity—using only a single sample per period—while accommodating non-stationary arrivals and initially unknown reward functions, advancing beyond the limitations of prior single-sample and episodic models.

Additional discussion of the further related work can be found in \Cref{apx:related work}.

\section{Preliminaries}
\label{sec:prelim}
We consider an online resource allocation problem over a finite horizon with $\numResource$ resources. Each resource $i \in [\numResource]$ has an initial capacity $\constraint_i\in \reals_{\ge0}$. The process takes place over $\numPeriod$ discrete time periods. At each period $t \in [\numPeriod]$, one single query arrives, which we denote as query $\query$. Each query $t$ is characterized by a \emph{random} resource consumption vector $\size_t=(a_{t,1},\dots,a_{t,\numResource}) \in \reals^\numResource_{\ge0}$, where $a_{t,i}$ represents the amount of resource $i$ that will be consumed to serve query $\query$ for all $i \in [\numResource]$, and a \emph{random} reward $\reward_t \in \reals_{\ge0}$ that denotes the amount of reward that can be collected by serving query~$\query$. 

We assume that for each period $t \in [\numPeriod]$, the pair $(\reward_t, \size_t)$ is drawn \emph{independently} from a nonstationary distribution denoted by $G_t(\cdot)$. Furthermore, we make the following structural assumption over the distribution $G_t(\cdot)$: queries belong to a finite number of types, where the resource consumption size is fixed for each query type, while the reward is continuous. 
For query $\query$, we define its type as $\type_t$, which is drawn \emph{independently} from a nonstationary distribution denoted by $\typedf_t(\cdot)$. For each $t \in [\numPeriod]$, the consumption vector $\size_t$ takes value in a finite set $\sizeset=\{\ba_1,\dots,\ba_\numSize\}$. When query $\query$ is of type $\type$, its consumption is given by $\size_t = \ba_j$. We denote $\typedf_t(j)=\prob{\size_t=\ba_j}$ for each $\type \in [\numSize]$ and each period $t$. Conditional on the query type—equivalently, given $\size_t = \ba_j$—the reward $\reward_t$ is independently and identically distributed according to a distribution $\cdf_j(\cdot)$. Regarding the distribution $\cdf_j(\cdot)$, we make the following assumption:

\begin{assumption}\label{assump:reward}
For each $\type \in [\numSize]$, the distribution function $\cdf_j(\cdot)$ has a density function $\pdf_j(\cdot)$ supported on the interval $[\lowreward_j, \uppreward_j]$ where $0<\lowreward_j<\uppreward_j<\infty$. Also, there exists two constants $0<\alpha<\beta<\infty$ such that for each $\type \in [\numSize]$ and each $r \in [\lowreward_j, \uppreward_j]$, it holds that
\begin{align*}
    \alpha\leq \pdf_j(r) \leq \beta.
\end{align*}
\end{assumption}

After query $\query$ arrives and its associated values $(\reward_t,\size_t)$ are revealed, the decision maker has to decide immediately and irrevocably whether or not to serve query $\query$, according to an online policy. Note that query $\query$ can only be served if for every resource $i$, the remaining capacity is at least $a_{t,i}$. The decision maker's objective is to maximize the total collected reward subject to the resource capacity constraint. 

For any online policy $\pi$, we define the decision variables as $\{\piDecision_t\}_{t=1}^\numPeriod$, where $\piDecision_t$ is a binary variable indicating whether query $\query$ is served. A policy $\pi$ is feasible if for every $t$, the decision $\piDecision_t$ depends solely on $\cdf_{\type_t}(\cdot)$ and previous instance $\{(\reward_s, \size_s)\}_{s=1}^t$, and if the following constraint is satisfied:
\begin{align*}
    \sum\nolimits_{t \in [\numPeriod]} a_{t,i} \cdot \piDecision_t \leq \constraint_i.
\end{align*}

The total reward of a policy $\pi$ is given by $V^{\pi}_{\bconstraint}(\instance) = \sum\nolimits_{t \in [\numPeriod]} \reward_t \cdot \piDecision_t$, where $\instance = \{(\reward_t, \size_t)\}_{t=1}^\numPeriod$ denotes the sample path of the problem instance and $\bconstraint=(\constraint_{1}, \cdots, \constraint_{\numResource})$.

\xhdr{Reward-Observed Samples versus Type-Only Samples.} We assume that both the query type distribution $\typedf_t(\cdot)$ and the conditional reward distribution $\cdf_j(\cdot)$ are unknown and must be learned. In each time period $t$, we have access to only one single sample from historical observation whose type $\stype_t$ is drawn independently from $\typedf_t$. We assume that \emph{reward-observed samples} include both information of arrival type $\stype_t$ and its corresponding reward $\sreward_t$ drawn from $\cdf_{\stype_t}$. In contrast, for \emph{type-only samples}, we have no historical reward information. For every $t \in [\numPeriod]$, the historical samples $\stype_t$ and $\sreward_t$ are independent of the problem instance random variables $\type_t$ and $\reward_t$.

\xhdr{Regret Minimization and Fluid Relaxation Benchmark.}
In order to establish a benchmark of decision policy $\pi$, it is common practice to adopt the offline optimal policy—that is, the policy chosen when the decision maker has full prior knowledge of $(\reward_t, \size_t)$ for all $t \in [\numPeriod]$. We denote the corresponding decisions as $\{\offDecision_t\}_{t=1}^\numPeriod$, which form the optimal solution to the following offline problem:
\begin{align}
    \label{eq:V_off} 
    \tag{$\Voff_{\bconstraint}(\instance)$}
    \arraycolsep=5.4pt\def\arraystretch{1.7}
    \begin{array}{llll}
    \max\limits_{\xbf} ~ &
    \displaystyle
    \sum\nolimits_{t\in[\numPeriod]} \reward_t \cdot x_t
    &
    \text{s.t.}
    \\
    &
    \displaystyle\sum\nolimits_{t\in[T]} a_{t,i} \cdot x_t \leq \constraint_i
    & i\in[\numResource]
    \\
    & x_t \in \{0,1\} & t \in [\numPeriod]
    \end{array}
\end{align}

We define the performance loss as \emph{regret}, which is the expected gap between the objective value of offline optimum and our policy $\pi$: 
\begin{equation*}
    \regret(\pi) := \expect[\instance \sim G]{\Voff_{\bconstraint}(\instance)} - \expect[\instance \sim G]{V_{\bconstraint}^{\pi}(\instance)}
\end{equation*}

However, in our setting, computing this offline optimal policy is challenging due to the stochastic nature of the problem instance $\instance$. To address this difficulty, we introduce its fluid relaxation. Specifically, for each query type $\type \in [\numSize]$, let $\numQuery_j$ represent the number of arriving queries with $\size_t = \ba_j$ and with $\reward_t$ drawn from $\cdf_j$ in the problem instance. 
To get rid of the dependence of the problem instance, we take the expectation of $\numQuery_j$ over the instance and have the following \emph{fluid} relaxation formulation:
\begin{align}
    \label{eq:V_fluid} 
    \tag{$\Vfluid_{\bconstraint}$}
    \arraycolsep=5.4pt\def\arraystretch{1.7}
    \begin{array}{llll}
    \max\limits_{\xbf} ~ &
    \displaystyle
    \sum\nolimits_{j\in[\numSize]} \expect[\instance \sim G]{\numQuery_j} \cdot \expect[\reward \sim \cdf_j]{\reward \cdot x_j(\reward)}
    &
    \text{s.t.}
    \\
    &
    \displaystyle\sum\nolimits_{j\in[\numSize]} \expect[\instance \sim G]{\numQuery_j} \cdot a_{j,i} \cdot \expect[\reward \sim \cdf_j]{x_j(\reward)} \leq \constraint_i
    & i\in[\numResource]
    \\
    & x_j(\reward) \in [0,1] & j \in [\numSize], \reward \in [\lowreward_j, \uppreward_j]
    \end{array}
\end{align}
where $\expect[\instance \sim G]{\numQuery_j} = \sum\nolimits_{t\in[\numPeriod]} \typedf_t(j)$. It's trivial that the relaxation imply an upper bound of the offline optimum \ref{eq:V_off}, as formalized in the following lemma:
\begin{lemma} \label{le:Vfluid_Voff}
    It holds that $\Vfluid_{\bconstraint} \geq \expect[\instance \sim G]{\Voff_{\bconstraint}(\instance)}$.
\end{lemma}

Therefore the \emph{regret} can be upper bounded by the expected gap between $\Vfluid_{\bconstraint}$ and $V_{\bconstraint}^{\pi}(\instance)$:
\begin{equation*}
    \regret(\pi) \leq \Vfluid_{\bconstraint} - \expect[\instance \sim G]{V_{\bconstraint}^{\pi}(\instance)}
\end{equation*}

In \ref{eq:V_fluid}, $\expect[\reward \sim \cdf_j]{x_j(\reward)}$ can be interpreted as the probability to serve type $\type$ queries dependent on reward $\reward$. We denote $\{x_j^*(\reward), \forall j \in [\numSize], \forall \reward \in [\lowreward_j, \uppreward_j]\}$ as an optimal solution to \ref{eq:V_fluid}. We have the following lemma to show the threshold property of the optimal solution:
\begin{lemma} \label{le:threshold_property}
    For an optimal solution $\{x_j^*(\reward), \forall j \in [\numSize], \forall \reward \in [\lowreward_j, \uppreward_j]\}$ to \ref{eq:V_fluid}, there exists a set of threshold $\{\kappa_j\}_{j=1}^{\numSize}$ such that it is optimal to set $x_j^*(\reward)=1$ if and only if $\reward \geq \kappa_j$, and $x_j^*(\reward)=0$ if and only if $\reward < \kappa_j$, for any $j \in [\numSize]$.
\end{lemma}

Consequently, following \Cref{le:threshold_property}, the \emph{fluid} relaxation problem can be equivalently rewritten as:
\begin{align}
    \label{eq:V_fluid2} 
    \tag{$\bVfluid_{\bconstraint}$}
    \arraycolsep=5.4pt\def\arraystretch{1.7}
    \begin{array}{llll}
    \max\limits_{\bq} ~ &
    \displaystyle
    \sum\nolimits_{j\in[\numSize]} \expect[\instance \sim G]{\numQuery_j} \cdot \int_{1-q_j}^1 \cdfInverse_j(u)\, \dd u
    &
    \text{s.t.}
    \\
    &
    \displaystyle\sum\nolimits_{j\in[\numSize]} \expect[\instance \sim G]{\numQuery_j} \cdot a_{j,i} \cdot q_j \leq \constraint_i
    & i\in[\numResource]
    \\
    & q_j \in [0,1] & j \in [\numSize]
    \end{array}
\end{align}
Here the decision variable $q_j$ represents the probability of serving a query of type $\type$. We denote $\{q_j^*\}_{\type=1}^\numSize$ as one optimal solution to \ref{eq:V_fluid2}. To minimize our \emph{regret},  we would ideally use this solution to design a feasible online policy.
However, the solution of fluid relaxation depends on the unknown distribution function $\{\typedf_t(\cdot)\}_{t=1}^\numPeriod$ and $\{\cdf_j(\cdot)\}_{\type=1}^\numSize$. Consequently, the theoretically optimal service probabilities $\{q_j^*\}_{\type=1}^\numSize$ cannot be directly implemented to derive a feasible online policy. Instead, we must adopt a data-driven approach that relies on historical observations to obtain reliable high-probability estimates for both $\expect[\instance \sim G]{\numQuery_j}$ and $\cdf_j$ for every $\type \in [\numSize]$. We will specify the estimation methods in \Cref{subsec:estimation}.

\section{Type-Dependent Quantile-Based Policy Framework}
\label{sec:meta algorithm}
In this section, we introduce a type-dependent quantile-based policy for the decision maker. The core idea of our approach is to translate a target service probability for each query type into a dynamic reward threshold, which is then used for make real-time decisions. This framework effectively decouples the problem: determining the optimal service probabilities $q_j$ becomes a fluid optimization task, while enforcing these probabilities online is accomplished via a carefully calibrated threshold rule.

We denote $q_j^{\pi}$ as the probability that query $\query$ will be served under policy $\pi$. Conceptually, this service probability serves as a key mechanism for balancing reward accumulation against resource consumption. Given the target service probability $\{q_j^{\pi}\}_{j=1}^\numSize$ and the estimated reward cumulative distribution function $\{\estcdf_j(\cdot)\}_{\type=1}^\numSize$ for each type $\type$, we can calculate a corresponding reward threshold $M(\estcdf_{\type}, q_{\type}^{\pi})$ to decide whether to accept or reject the arriving query based on the observed reward. This threshold is defined as the quantile of the estimated distribution above which a fraction $q_j^{\pi}$ of the rewards lie.

Upon arrival of a type-$\type$ query with reward, our policy accepts it if and only if the reward meets or exceeds the threshold and sufficient resources remain. Intuitively, by accepting only rewards above the ($1 - q_j^{\pi}$)-th quantile, the empirical service rate for type $\type$ converges to the target $q_j^{\pi}$, provided our distribution estimate is accurate.
Our proposed meta-policy is formalized in Subroutine \ref{alg:meta}. It operates online and takes an exogenous threshold $M(\estcdf_{\type_t}, q_{\type_t}^{\pi})$ as an key input at each time period $t$. The specific choice of this threshold distinguishes between different settings, which will be detailed in the following \Cref{sec:offline setting}, \ref{sec:online setting} and \ref{sec:poly-log regret} respectively.

\begin{myprocedure}
\caption{\textsc{Meta Quantile-based Policy}}
\label{alg:meta}
\SetKwInput{KwInput}{Input}
\SetKwInput{KwOutput}{Output}
\KwInput{ 
Arrival type $\type_t$ and reward $\reward_t$; 
Quantile-based threshold $M(\estcdf_{\type_t}, q_{\type_t}^{\pi})$;
Current consumption $\size_{\type_t}$;
Remaining budget $\constraint_{t,i}$ for every $i \in [\numResource]$.
}
\KwOutput{
Decision: ACCEPT or REJECT the query;
Updated budgets $\constraint_{t+1,i}$ for all $i \in [\numResource]$.
}
\If{$\reward_t \geq M(\estcdf_{\type_t}, q_{\type_t}^{\pi})$ \textbf{and} $\constraint_{t,i} \geq a_{\type_t,i}, \forall i \in [\numResource]$}{
    ACCEPT the query and record $\consumption_i = a_{\type_t,i}$ for every $i \in [\numResource]$\;
}
\Else{
    REJECT the query and record $\consumption_i = 0$ for every $i \in [\numResource]$\;
}
Update the remaining budget $\constraint_{t+1,i} = \constraint_{t,i} - \consumption_i$ for every $i \in [\numResource]$.
\end{myprocedure}

The elegance of this meta quantile-based policy lies in its modular design. The complexity of learning reward distributions and determining appropriate service probabilities is abstracted into the inputs $\estcdf_j$ and $q_j^{\pi}$. In turn, the online decision rule reduces to a straightforward threshold comparison, making the policy highly practical for real-time deployment in latency-sensitive systems.
Overall, our work represents a paradigm shift from integrated, constraint-driven learning to a modular, estimation-first architecture, significantly advancing both analytical tractability and empirical adaptability in online resource allocation.

\subsection{$\Tilde{O}(\sqrt{T})$ Regret by Static and Partial Adaptive Thresholds}
Our quantile-based framework achieves $\Tilde{O}(\sqrt{T})$ regret through two distinct approaches: (i) static thresholds derived from samples with observed rewards, and (ii) partially adaptive thresholds leveraging only query-type information. Both policies operate within the unified modular meta quantile-based policy (Subroutine \ref{alg:meta}), which decouples distribution estimation from real-time decision-making via type-dependent reward thresholds.

In \Cref{sec:offline setting}, we assume that historical data contain both query types and corresponding reward values. Before the online process begins, we use the single sample per period to construct kernel-based estimators $\estcdf_j$ for reward distribution of each arrival type. We then formulate an estimated fluid relaxation \ref{eq:V_est_offline} that replaces unknown expected arrivals $\expect[\instance \sim G]{\numQuery_j}$ with historical counts $\snumQuery_j$ and true distributions $\cdf_j$ with their estimates $\estcdf_j$. Solving this estimation problem via its Lagrangian function yields target service probabilities $\hq_j$ and we can calculate the corresponding acceptance thresholds $M(\estcdf_{\type}, q_{\type}^{\pi})$ by setting $q_{\type}^{\pi}$ as $\hq_j$. These thresholds are passed to Subroutine \ref{alg:meta} for real-time decisions. Although estimation errors cause the implemented solution to deviate from the true fluid optimum, we rigorously bound the resulting regret by analyzing the performance gap and the penalty due to possible constraint violations. \Cref{th:offline} shows that our static threshold policy (\Cref{alg:offline}) attains a regret of $O(\numResource \log\numPeriod \sqrt{\numSize \numPeriod})$.

Our novel approach fundamentally shifts from dual-based frameworks, achieving superior theoretical performance through a more direct sample-coupled analysis and decoupled design. Prior work on exponential pricing \citep{GSW-25} relies heavily on no-regret properties with respect to the zero price vector, requiring intricate concentration analysis to control estimation errors in prefix consumption. Similarly, robust pacing algorithms \citep{BKMSW-23} build upon online convex optimization frameworks where regret bounds translate to solution quality guarantees. In contrast, our method leverages coupling arguments between samples and realizations.
This decoupled design not only enhances practical deployability but also contributes to a tighter regret bound. To be specific, our static thresholds method achieves a $\tilde{O}(\sqrt{\numPeriod})$ regret bound, which outperforms the $\tilde{O}(\numPeriod^{5/6})$ rate in \citep{GSW-25} when resource scales linearly with time horizon and matched the $\tilde{O}(\sqrt{\numPeriod})$ bound in \citep{BKMSW-23} with single-resource setting.

A further distinction lies in decision granularity. Conventional dual-based methods employ a single global price vector coupling all resources and types, obscuring per-query logic and entangling decisions with global state. While our framework enables transparent, type-specific decisions: each query type is managed independently by its own quantile threshold. This fundamental shift not only eliminates the need for cumbersome global estimation but also introduces greater flexibility, as each query type can be processed independently through its own quantile-based threshold without interference from others.

In \Cref{sec:online setting}, we consider a more difficult setting where historical samples contain only query types and no rewards. We first establish an impossibility result (\Cref{th:impossible_sublinear_regret}): without further assumptions, no online policy can achieve sublinear regret, as illustrated by a counterexample where non-stationary arrivals force a linear regret $\Omega(\numPeriod)$. To enable learning, we introduce a mild minimum-arrival-probability assumption (\Cref{assump:MAP}), guaranteeing each type appears with probability at least $\gamma > 0$ per period. Under this condition, we design a partial adaptive threshold scheme. Starting with a uniform prior, each time a query of type $\type$ arrives, we update the estimator $\estcdf_{j,t}$ using the newly observed reward. At period $t$, we solve an online version of the estimated relaxation \ref{eq:V_est_online} using current estimates $\estcdf_{j,t}$ and historical counts $\snumQuery_j$, producing time-varying target service probabilities $q_{j,t}$ and corresponding thresholds $M(\estcdf_{\type,t}, q_{\type,t})$. The resulting online policy (\Cref{alg:online}) continuously refines distribution estimates while respecting resource constraints. Remarkably, \Cref{th:online} confirms that the same $O(\numResource \log\numPeriod \sqrt{\numSize \numPeriod})$ regret bound holds—demonstrating that online reward learning need not compromise theoretical performance when minimal arrival regularity is ensured.
 
\subsection{Logarithmic Regret by Fully Adaptive Thresholds}
To surpass the $\sqrt{\numPeriod}$ barrier, in \Cref{sec:poly-log regret} we introduce a fully adaptive resolving policy that re-optimizes the remaining problem at each decision epoch and incorporates a careful rounding step. This method is also designed for type-only samples under minimum-arrival-probability condition (\Cref{assump:MAP2}).
The core idea is to move from a static or partially adaptive fluid relaxation to a semi-fluid relaxation that explicitly accounts for the remaining horizon and current resource capacities. At the beginning of each period $t$, we compute the estimated future arrivals $\tsnumQuery_{j,t}$ from historical samples and current reward estimates $\estcdf_{j,t}$. Using these, we formulate a per-period estimated problem \ref{eq:V_est_log_t} with remaining capacities $\brconstraint_t$. Solving this problem via Lagrangian function yields candidate service probabilities $\hq_{j,t}$. To ensure robustness, we introduce a rounding procedure when setting the acceptance threshold:
(i) If $\hq_{\type_t,t}$ is very high (above $1-2\kappa (\frac{\log\numPeriod}{\sqrt{\numPeriod-t+1}}+\frac{\log\numPeriod}{\sqrt{t}})$), we set the threshold to the lower bound $\lowreward_{\type_t}$, effectively accepting all queries of that type;
(ii) If $\hq_{\type_t,t}$ is very low (below $2\kappa (\frac{\log \numPeriod}{\sqrt{\numPeriod-t+1}}+\frac{\log \numPeriod}{\sqrt{t}})$), we set the threshold to $\uppreward_{\type_t}+1$, effectively rejecting all queries of that type;
(iii) In the intermediate range, we use the quantile-based threshold $\estcdfInverse_{\type_t,t}(1-\hq_{\type_t,t})$.

This rounding policy ensures that we could always constructs a feasible solution to our benchmark. By analyzing the instantaneous regret at each step and aggregating over the horizon, we prove \Cref{th:log} that the fully adaptive threshold policy (\Cref{alg:log}) achieves a regret of $O((\log\numPeriod)^3)$. This represents an exponential improvement over the typical $\Tilde{O}(\sqrt{T})$ rates in prior dual-based literature.
The decoupled design also enables that we can employ aggressive online learning methods of distributions because reward estimation is separated from resource budgeting, while in dual‑based approaches, the interleaving of dual updates and reward estimation makes it difficult to control the error propagation needed for logarithmic regret.
To conclude, our poly‑logarithmic regret bound bridges a significant theoretical gap in online resource allocation. It demonstrates that, even with extreme data scarcity (one sample per period) and no prior reward knowledge, near‑optimal performance is attainable with only logarithmic growth in regret.

\subsection{Estimation of Arrival Distributions} \label{subsec:estimation}
In this subsection, we present standard methods for estimating query arrival numbers and reward distributions, which serve as direct inputs to our modular framework.

\xhdr{Estimation of Query Numbers.}
Given the non-stationary nature of the environment, estimating query type distributions is not meaningful. Instead, we focus on estimating the total number of query arrivals for each type over the entire time horizon. We now specify our approach:
For each period $t$, we have one single historical sample $\stype_t \sim \typedf_t$. Since the query types are finite, i.e. $\stype_t \in [\numSize]$, we denote the number of type $\type$ query arrivals from historical observation as $\snumQuery_j = \sum\nolimits_{t \in [\numPeriod]} \indicator{\stype_t=j}$. It follows directly that $\sum\nolimits_{j \in [\numSize]} \snumQuery_j = \numPeriod$ as we have exactly $\numPeriod$ samples in total. We make the following assumption:
\begin{assumption}\label{assump:d}
For any historical instance with $\bsnumQuery = (\snumQuery_{1}, \cdots, \snumQuery_{\numSize})$, at least one query of each type $\type$ arrives, i.e.,
\begin{equation*}
    \prob{\snumQuery_j \geq 1} = 1, \forall j \in [\numSize].
\end{equation*}
\end{assumption}

For a given problem instance $\instance$, the actual number of type $\type$ query arrivals is $\numQuery_j = \sum\nolimits_{t \in [\numPeriod]} \indicator{\type_t=j}$. We have the following lemma to show that $\snumQuery_j$ serves as a reliable estimate for $\expect[\instance \sim G]{\numQuery_j}$. We offer detailed discussion in \Cref{apx:est_arrival-distribution}. 
\begin{lemma} \label{le:est_dj}
Define $\mu_j = \expect[\instance \sim G]{\numQuery_j}$, with probability at least $1-\delta$, we have 
\begin{equation*}
    \sum\nolimits_{j \in [\numSize]} |\snumQuery_j - \mu_j| \leq \sqrt{2\numSize \numPeriod \log(2/\delta)}, \quad \sum\nolimits_{j \in [\numSize]} \frac{|\snumQuery_j - \mu_j|^2}{\mu_j} \leq \numSize \left(\log(2/\delta)\right)^2
\end{equation*}
\end{lemma}

\xhdr{Estimation of Reward Distribution.}
Our quantile-based framework is compatible with general reward distribution estimation methods. In our analysis, we adopt a kernel-based estimation approach:
Let kernel probability density function $k(x)$ satisfies the following properties: (i) $k(x)$ is non-negative symmetrical density function; (ii) $k(x)$ has compact support on [-1,1], i.e., $k(x)=0$ when $|u|>1$;  (iii) $\int k(x)\, \dd x = 1, \int xk(x)\, \dd x = 0, \int |x|k(x)\, \dd x < +\infty$. We define $K(x) = \int_{-\infty}^x k(u)\, \dd u$ as the cumulative distribution function of $k(x)$. Therefore we know $K(x)$ is symmetric: $K(-x)=1-K(x)$ and $K(0)=\frac{1}{2}$.

Assume that we have $\sample$ i.i.d. samples $\{X_1, \cdots, X_{\sample}\}$ drawn from a reward distribution function $\cdf(\cdot)$ and $\cdf(\cdot)$ is supported on $[a,b]$ where $0<a<b<\infty$. The probability density function $\pdf(\cdot)$ satisfies that $\forall x \in [a,b], 0<\alpha \leq \pdf(x) \leq \beta < \infty$. Define the kernel estimation as following:
\begin{equation} \label{eq:F_kernel_estimation}
    \estcdf(x) = \frac{1}{\sample} \sum_{i=1}^{\sample} K\left(\frac{x - X_i}{h_{\sample}}\right)
\end{equation}
where $h_{\sample} > 0$ is the bandwidth parameter to be specified. Then we have the following lemma, the proof of which is detailed in \Cref{apx:est_arrival-distribution}.
\begin{lemma} 
\label{le:est_Fj}
Given $\sample$ i.i.d. samples $\{X_1, \cdots, X_{\sample}\}$ drawn from a reward distribution function $\cdf(\cdot)$, set $h_{\sample} = \sample^{-1/2}$, $\estcdf(x)$ defined in (\ref{eq:F_kernel_estimation}), with probability at least $1- \delta$, the uniform estimation error satisfies
\begin{equation*}
    \sup_{x \in [a,b]} |\cdf(x) - \estcdf(x)| \leq O\left(\sqrt{\frac{\log(1/\delta)}{\sample}}\right)
\end{equation*}
\end{lemma}

\section{Static Thresholds with Reward-Observed Samples}
\label{sec:offline setting}
In this section, we assume that the historical samples (reward-observed samples) also contain the corresponding reward values $\sreward_t$, each drawn from $\cdf_{\stype_t}$. This allows us to compute, with high probability, an estimator for the reward distribution function of each type $\type$ prior to the decision process.
Specifically, we utilize the $\snumQuery_j$ historical samples associated with each type $\type$ to construct an estimate of the reward distribution $\cdf_j(\cdot)$ with kernel method stated in \Cref{subsec:estimation}.

\xhdr{Threshold Computation and Algorithm Design:}
We now develop the estimator threshold that will be used in Subroutine \ref{alg:meta}. Given  historical samples $\bsnumQuery = (\snumQuery_{1}, \cdots, \snumQuery_{\numSize})$, we define the estimation problem of \ref{eq:V_fluid2}:
\begin{align}
    \label{eq:V_est_offline}
    \tag{$\hV_{\bconstraint}(\bsnumQuery)$}
    \arraycolsep=5.4pt\def\arraystretch{1.7}
    \begin{array}{llll}
    \max\limits_{\bq} ~ &
    \displaystyle
    \sum\nolimits_{j\in[\numSize]} \snumQuery_j \cdot \int_{1-q_j}^1 \estcdfInverse_j(u)\, \dd u 
    &
    \text{s.t.}
    \\
    &
    \displaystyle\sum\nolimits_{j\in[\numSize]} \snumQuery_j \cdot a_{j,i} \cdot q_j \leq \constraint_i
    & i\in[\numResource]
    \\
    & q_j \in [0,1] & j \in [\numSize]
    \end{array}
\end{align}

We can solve the estimation problem \ref{eq:V_est_offline} and obtain its optimal solution:
\begin{equation} \label{eq:q_offline}
    \hq_j(\hblambda) = 1 - \estcdf_j\left(\sum\nolimits_{i \in [\numResource]} \hlambda_i a_{j,i}\right)
\end{equation}
where $\hblambda$ is the optimal dual variable in the Lagragian function of \ref{eq:V_est_offline}.

To be specific, We summarize Subroutine \ref{subroutine} to serve as a solver to the estimation problem.
\begin{myprocedure}
\caption{\textsc{Empirical Quantile Thresholds Solver}}
\label{subroutine}
\SetKwInput{KwInput}{Input}
\SetKwInput{KwOutput}{Output}
\SetKwBlock{KwInit}{Processing:}{}
\KwInput{ 
Estimated query number $\{\snumQuery_j\}_{j=1}^{\numSize}$; Estimated reward distribution function $\{\estcdf_j(\cdot)\}_{j=1}^{\numSize}$; Resource consumption vectors $\ba_j$ for every $j \in [\numSize]$; Initial resource capacity $\bconstraint$.
}
\KwOutput{
Optimal solution $\{\hq_j(\hblambda)\}_{j=1}^{\numSize}$.
}
Formulize the Lagragian function as $\hL(\bq,\blambda) = \sum\nolimits_{j\in[\numSize]} \snumQuery_j \int_{1-q_j}^1 \estcdfInverse_j(u)\, \dd u - \sum\nolimits_{i \in [\numResource]} \lambda_i (\sum\nolimits_{j\in[\numSize]} \snumQuery_j a_{j,i} q_j - \constraint_i)$\;
Solve the dual variable as $\hblambda = \argmin_{\blambda} \left(\max_{\bq} \hL(\bq,\blambda)\right)$\;
Compute the target service probability as $\hq_j(\hblambda) = 1 - \estcdf_j(\sum\nolimits_{i \in [\numResource]} \hlambda_i a_{j,i})$.
\end{myprocedure}

In the context of our policy, the most straightforward way to implement the service probability is to set $q_j^{\pi} := \hq_j(\hblambda)$. Based on the preceding analysis, we can calculate the estimator threshold as following:
\begin{equation*} \label{eq:threshold_offline}
    M(\estcdf_{\type}, q_{\type}^{\pi}) := \estcdfInverse_{\type}(1-q_{\type}^{\pi})
\end{equation*}

Our static threshold policy is therefore formalized in \Cref{alg:offline}. 

\begin{algorithm}
\caption{{\STP}}\label{alg:offline}
\SetAlgoLined
\SetKwInput{KwInput}{Input}
\SetKwBlock{KwInit}{Initialization:}{}
\KwInput{
Historical samples: $\{(\hat{\type}_t, \hat{\reward}_t)\}_{t=1}^T$;
Resource capacities $\constraint_i \in \reals_{\ge0}$ for every $i \in [\numResource]$;
Resource consumption vectors $\ba_j$ for every type $j \in [\numSize]$.
}
Initialize remaining capacities $\constraint_{1,i} = \constraint_i$ for every $i \in [\numResource]$\;
Count the number of samples for each type $\type$ as $\snumQuery_j$\;
Estimate the distribution of type $j$ reward $\estcdf_j$ with $\snumQuery_j$ samples using kernel estimation\;
Construct the estimated relaxation problem as \ref{eq:V_est_offline}\;
Follow Subroutine \ref{subroutine} to compute service probabilities $q_{\type}^{\pi}$ as (\ref{eq:q_offline}) for every $j \in [\numSize]$\;
Compute the threshold $ M(\estcdf_{\type}, q_{\type}^{\pi}) = \estcdfInverse_{\type}(1-q_{\type}^{\pi})$ for every $j \in [\numSize]$.

\For{$t = 1, \cdots, T$}{
Observe arrival type $\type_t$ and reward $\reward_t$\;
Call Subroutine \ref{alg:meta} with input $(\type_t, \reward_t,  M(\estcdf_{\type_t}, q_{\type_t}^{\pi}), \size_{\type_t})$ and $\constraint_{t,i}$ for every $i \in [\numResource]$. 
}
\end{algorithm}

The algorithm operates in two phases: an offline pre-processing phase followed by an online decision phase. In the offline phase, the algorithm initializes the resource capacities and uses historical data to estimate the query arrival numbers and reward distributions. It then constructs and solves an estimated relaxation problem to obtain target service probabilities for each query type as Subroutine \ref{subroutine}. These probabilities are subsequently converted into type-specific reward thresholds. In the online phase, for each arriving query, the policy compares its observed reward against the pre-computed threshold. The final accept/reject decision is made by our meta quantile-based policy (Subroutine \ref{alg:meta}), which also accounts for the current remaining resource budget.

We state our main result in the following theorem:
\begin{theorem} 
\label{th:offline}
With reward-observed samples, under \Cref{assump:reward} and \ref{assump:d}, the regret of 
{\STP} (\Cref{alg:offline}) is at most
\begin{equation*}
    (\uppreward + \uppreward \numResource \amax)\left(\frac{4\beta + 2\beta k_1 }{\alpha} + 2 + \uppreward\right) \log\numPeriod \sqrt{\numSize \numPeriod} = O(\numResource \log\numPeriod \sqrt{\numSize \numPeriod})
\end{equation*}
where $\numResource$ represents the number of resource constraints, $\numSize$ represents the number of arrival query types and $\uppreward = \max_j \uppreward_j, \amax = \max_{j,i} a_{j,i} , \alpha, \beta, k_1$ are all constants.
\end{theorem}

\xhdr{Performance Analysis:}
We now outline the analysis of the performance loss incurred by our approach. When accurate estimators are available for both $\expect[\instance \sim G]{\numQuery_j}$ and $\cdf_j(\cdot)$, estimated solution $\{\hq_j(\hblambda)\}_{j=1}^\numSize$ can be used to approximate true primal optimum. However, in practice, estimation errors inevitably introduce a performance gap between our policy and theoretical benchmark. To assess the robustness of our approach, we now examine the optimal benchmark solution and demonstrate that the resulting regret remains bounded.
Denote $\mu_j$ as the expectation of $\numQuery_j$. We define the Lagrangian function of primal problem \ref{eq:V_fluid2}:
\begin{equation*}
    \begin{aligned}
        L(\bq,\blambda) & = \sum\nolimits_{j\in[\numSize]} \expect[\instance \sim G]{\numQuery_j} \int_{1-q_j}^1 \cdfInverse_j(u)\, \dd u - \sum\nolimits_{i \in [\numResource]} \lambda_i (\sum\nolimits_{j\in[\numSize]} \expect[\instance \sim G]{\numQuery_j} a_{j,i} q_j - \constraint_i)\\
        & = \sum\nolimits_{j\in[\numSize]} \mu_j \int_{1-q_j}^1 \cdfInverse_j(u)\, \dd u - \sum\nolimits_{i \in [\numResource]} \lambda_i (\sum\nolimits_{j\in[\numSize]} \mu_j a_{j,i} q_j - \constraint_i)
    \end{aligned}
\end{equation*}
and its optimal solution can be written as
\begin{equation} \label{eq:q_optimal}
    q_j^*(\blambda^*) = 1 - \cdf_j\left(\sum\nolimits_{i \in [\numResource]} \lambda_i^* a_{j,i}\right)
\end{equation}
where $\blambda^*$ is the optimal dual variable of the Lagragian function.

We can demonstrate that the performance loss can be expressed as a function of the difference between the optimal benchmark solution $\{q_j^*(\blambda^*)\}_{j=1}^\numSize$ and the solution employed by our algorithm.

It is important to note that, due to the budget constraints inherent in the problem, directly applying the estimated solution in our algorithm can also lead to practical constraint violations. This arises because the constraints in the primal problem (based on true parameters) differ from those in the estimation problem (based on sampled data). To address this issue, we formally quantify the penalty of such constraint violations through the term $V(\Hat{\bq}(\hblambda))$, defined in \Cref{eq:Violation_offline}. This term captures the expected excess resource consumption incurred by implementing potentially infeasible estimated solution. Importantly, we can demonstrate that the impact of this constraint violation is manageable within our framework. Specifically, it contributes only a sublinear term to the overall regret.

To conclude, the result is formally stated in the proof of \Cref{th:offline} (see \Cref{apx:offline}), which guarantees the near-optimal performance of our static threshold policy despite the inherent estimation error. The theorem essentially shows that although perfect feasibility cannot be ensured when operating with estimated parameters, the resulting performance loss scales favorably with the sample size and does not dominate the long-term average reward.

\section{Partial Adaptive Thresholds with Type-Only Samples}
\label{sec:online setting}
When we have type-only samples, historical reward information is unavailable, necessitating online learning of the reward distributions. Consequently, our goal is to sequentially update our estimate of reward distribution function and use the evolving solution to approximate the primal optimal solution in each step.

We commence by presenting a theorem that establishes an impossibility result:
\begin{theorem} \label{th:impossible_sublinear_regret}
    Under \Cref{assump:reward} and \ref{assump:d}, no online policy can beat the $\Omega(\numPeriod)$ regret lower bound with type-only samples.
\end{theorem}

We prove \Cref{th:impossible_sublinear_regret} by providing the following counterexample:

\begin{example} \label{example1}
    Consider a setting with one resource constraint $m=1$ and two query types $n=2$. The resource capacity $C=\frac{T}{2}$ and each query consumes one unit of resource, i.e., $a_1=a_2=1$. During the first half of the time horizon, only type $1$ queries arrive, so $\typedf_t(1)=1, \typedf_t(2)=0$ for $t = 1,2,\cdots,\frac{T}{2}$. In the second half, only type $2$ queries arrive, so $\typedf_t(1)=0, \typedf_t(2)=1$ for $t = \frac{T}{2}+1,\cdots,T$.
    
    We consider two scenarios: In Scenario 1, the reward distribution for type $1$ is uniform on $[1,2]$ (denoted $U[1,2]$), and for type $2$ is $U[0,1]$. In Scenario 2, type $1$ rewards follow $U[1,2]$, while type $2$ rewards follow $U[2,3]$.
    Thus, the benchmark policy (with full arrival information) accepts all high-reward queries: type $1$ in Scenario 1 and type $2$ in Scenario 2. Since the total resource capacity equals the number of arrivals for each high-reward type, the expected benchmark reward is $\frac{3T}{4}$ in Scenario 1 and $\frac{5T}{4}$ in Scenario 2.
    
    Let $T_1(\pi)$ and $T_2(\pi)$ denote the expected amount of resource consumed by policy $\pi$ during the first $\frac{T}{2}$ time periods in Scenario 1 and Scenario 2 respectively. Equivalently, $T_1(\pi)$ and $T_2(\pi)$ represent the expected numbers of queries accepted by policy $\pi$ in each scenario over that interval.
    Then the expected reward collected by policy $\pi$ in the two scenarios can be written as:
    \begin{equation*}
        \begin{aligned}
            & R_T^1(\pi) = \frac{3}{2}T_1(\pi) + \frac{1}{2}\left(\frac{T}{2}-T_1(\pi)\right) =  \frac{T}{4}+T_1(\pi)\\
            & R_T^2(\pi) = \frac{3}{2}T_2(\pi) + \frac{5}{2}\left(\frac{T}{2}-T_2(\pi)\right) = \frac{5T}{4}-T_2(\pi)
        \end{aligned}
    \end{equation*}
    
    Therefore, the regret of policy $\pi$ in Scenario 1 is $\frac{T}{2}-T_1(\pi)$, and in Scenario 2 it is $T_2(\pi)$. Moreover, because the policy $\pi$ can only depend on historical samples, its decisions during the first half must be identical in both scenarios; hence, we must have $T_1(\pi)=T_2(\pi)$. Consequently, we obtain
    \begin{equation*}
        \regret(\pi) \geq \max\{\frac{T}{2}-T_1(\pi), T_2(\pi)\} \geq \frac{T}{4} = \Omega(T)
    \end{equation*}
\end{example}

The example above illustrates that without knowledge of future arrivals, it is impossible to achieve sublinear regret in both scenarios simultaneously. If the policy conserves too much resource in the first half, then in Scenario 1 it misses the opportunity to accept profitable type 1 arrivals early on, which cannot be recovered. Conversely, if it consumes too much resource in the first half, then in Scenario 2 it lacks sufficient capacity to accept the profitable type 2 arrivals later.

Consequently, to enable online learning, we must impose a minimal arrival probability assumption:
\begin{assumption}\label{assump:MAP}
There exists a constant $\gamma > 0$ such that the type arrival probability $\typedf_t(j) \geq \gamma$, for every $j \in [\numSize]$ and $t \in [\numPeriod]$. 
\end{assumption}

\xhdr{Threshold Computation and Algorithm Design:}
We now develop the estimator threshold that will be used in Subroutine \ref{alg:meta}. The only difference from \Cref{sec:offline setting} is that we apply a \emph{partial adaptive} method to update our estimated reward distributions during the implementation of the problem instance.
Given  historical samples $\bsnumQuery = (\snumQuery_{1}, \cdots, \snumQuery_{\numSize})$, for each time period $t$, we define the online estimation problem:
\begin{align}
    \label{eq:V_est_online}
    \tag{$\hV_{t,\bconstraint}(\bsnumQuery)$}
    \arraycolsep=5.4pt\def\arraystretch{1.7}
    \begin{array}{llll}
    \max\limits_{\bq} ~ &
    \displaystyle
    \sum\nolimits_{j\in[\numSize]} \snumQuery_j \cdot \int_{1-q_j}^1 \estcdfInverse_{j,t}(u)\, \dd u 
    &
    \text{s.t.}
    \\
    &
    \displaystyle\sum\nolimits_{j\in[\numSize]} \snumQuery_j \cdot a_{j,i} \cdot q_j \leq \constraint_i
    & i\in[\numResource]
    \\
    & q_j \in [0,1] & j \in [\numSize]
    \end{array}
\end{align}
where $\estcdf_{j,t}$ is the estimator for $\cdf_j$ in time period $t$.
Then following the Subroutine \ref{subroutine} by substituting $\estcdf_j$ for $\estcdf_{j,t}$ at each time period, we can solve the optimal solution to \ref{eq:V_est_online}:
\begin{equation} \label{eq:q_online}
    q_{j,t}(\blambda_t) = 1 - \estcdf_{j,t}\left(\sum\nolimits_{i \in [\numResource]} \lambda_{t,i} a_{j,i}\right)
\end{equation}
where $\blambda_t$ is the optimal dual variable in Lagragian function of \ref{eq:V_est_online}.
Setting $q_{j,t}^{\pi} := q_{j,t}(\blambda_t)$ for each time period $t$, we can calculate the estimator threshold as following:
\begin{equation*} \label{eq:threshold_online}
    M(\estcdf_{\type,t}, q_{\type,t}^{\pi}) := \estcdfInverse_{\type,t}(1-q_{\type,t}^{\pi})
\end{equation*}

We now present our partial adaptive threshold policy in \Cref{alg:online}. 

\begin{algorithm}
\caption{{\PATP}}\label{alg:online}
\SetAlgoLined
\SetKwInput{KwInput}{Input}
\KwInput{
Historical samples: $\{\hat{\type}_t\}_{t=1}^T$;
Resource capacities $\constraint_i \in \reals_{\ge0}$ for every $i \in [\numResource]$;
Resource consumption vectors $\ba_j$ for every type $j \in [\numSize]$.
}
Initialize remaining capacities $\constraint_{1,i} = \constraint_i$ for every $i \in [\numResource]$\;
Count the number of samples for each type $\type$ as $\snumQuery_j$\;
Initialize distribution estimates $\estcdf_{j,0}$ as uniform distribution for all $j \in [\numSize]$\;

\For{$t = 1, \cdots, T$}{
Observe arrival type $\type_t$ and reward $\reward_t$\;
Update the estimation $\estcdf_{\type_t,t}$ using the new sample $\reward_t$; Set $\estcdf_{\type,t} = \estcdf_{\type,t-1}, \forall \type \neq \type_t$\; 
Construct the estimated relaxation problem as \ref{eq:V_est_online}\;
Follow Subroutine \ref{subroutine} to compute service probabilities $q_{\type,t}^{\pi}$ as (\ref{eq:q_online}) for every $j \in [\numSize]$\;
Compute current threshold $M(\estcdf_{\type_t,t}, q_{\type_t,t}^{\pi}) = \estcdfInverse_{\type_t,t}(1-q_{\type_t,t}^{\pi})$\;
Call Subroutine \ref{alg:meta} with input $(\type_t, \reward_t, M(\estcdf_{\type_t,t}, q_{\type_t,t}^{\pi}), \size_{\type_t})$ and $\constraint_{t,i}$ for every $i \in [\numResource]$.
}
\end{algorithm}

The policy begins by initializing the remaining resource capacities and estimating query arrival numbers from the input data, while initializing all reward distribution estimates as uniform. During the online execution, for each arriving query, we observe its type and reward value, and immediately use this new sample to update the reward distribution estimate for that specific type. We then reconstruct and solve the estimated relaxation problem based on the latest distributions to obtain updated optimal service probabilities for the current period. Finally, we compute a new adaptive threshold, which is passed to the meta quantile-based policy to guide the immediate decision. This process allows the thresholds to evolve in response to newly observed reward information.

The main result of this section is formalized as following:
\begin{theorem} \label{th:online}
With type-only samples, under \Cref{assump:reward}, \ref{assump:d} and \ref{assump:MAP}, the regret of {\PATP} (\Cref{alg:online}) is at most
\begin{equation*}
    (\uppreward + \uppreward \numResource \amax)\left(\frac{4\beta}{\alpha} + \frac{8\beta k_1}{\alpha \gamma} + \frac{16}{\gamma} + 2\uppreward\right) \log\numPeriod \sqrt{\numSize \numPeriod} = O(\numResource \log\numPeriod \sqrt{\numSize \numPeriod})
\end{equation*}
where $\numResource$ represents the number of resource constraints, $\numSize$ represents the number of arrival query types and $\uppreward = \max_j \uppreward_j, \amax = \max_{j,i} a_{j,i}, \alpha, \beta, \gamma, k_1$ are all constants.
\end{theorem}

This theorem demonstrates that with minimal arrival probability, despite the absence of prior reward knowledge, the policy achieves sublinear regret, effectively balancing exploration with exploitation while maintaining near-optimal resource utilization over time.

\xhdr{Performance Analysis:}
Regarding the regret of our partial adaptive threshold policy, it can still be expressed as a function of the difference between the optimal benchmark solution in (\ref{eq:q_optimal}) and the solution employed by our algorithm.
Similar to \Cref{sec:offline setting}, we also need to account for potential constraint violation. We quantify the impact of this infeasibility by $V(\Hat{\bq}(\blambda_t))$ defined in \Cref{eq:Violation_online}, which represents the penalty associated with constraint violation at step $t$.
By implementing online learning and decision-making procedure outlined in \Cref{alg:online}, we can manage these dual challenges of distributional learning and constraint satisfaction. The performance of this policy is characterized by a result analogous to the reward-observed sample case, formally presented in \Cref{th:online} with detailed proof in \Cref{apx:online}.

\section{Fully Adaptive Thresholds with Type-Only Samples}
\label{sec:poly-log regret}

In this section, we introduce a poly-logarithmic regret policy with fully adaptive threshold method using type-only samples. We consider the same setting as \Cref{sec:online setting} where the historical samples include no reward information. We state the following minimal-arrival-probability assumption:
\begin{assumption}\label{assump:MAP2}
There exists a constant $\gamma > 0$ such that the type arrival probability $\typedf_t(j) \geq \gamma$, for every $j \in [\numSize]$ and $t \in [\numPeriod]$. Here $\gamma$ is known to the decision maker.
\end{assumption}

Previous literature rigorously establishes that static policies, which solve the relaxation problem only once (from period $1$ to $\numPeriod$) and fix decisions thereafter, incur a regret lower bound $\Omega(\sqrt{\numPeriod})$ (see Theorem 3 in \citep{AG-19}), making logarithmic performance unattainable.
Consequently, adaptive resolving (from period $t$ to $\numPeriod$) at each time period $t$ is a necessary condition to circumvent the fundamental $\sqrt{\numPeriod}$ regret limitation and realize logarithmic-order guarantees, as validated by modern frameworks. 

In this paper, our static and partially adaptive policies rely on solutions computed based on the \emph{total initial resource capacity}. By contrast, the \emph{fully adaptive policy} studied in this section repeatedly re‑solves the problem using the \emph{remaining capacity} at each period, thereby enabling dynamic adjustment to real-time resource availability.

\xhdr{Algorithm Design:}
We give a general description of our \emph{fully adaptive threshold} approach. We define random variable for type $\type$ query arrivals from period $t$ to $\numPeriod$ as $\tsnumQuery_{j,t}$ in the historical samples. Denote $\brconstraint_t=(\rconstraint_{t,1}, \cdots, \rconstraint_{t,\numResource}) \in \reals^\numResource_{\ge0}$ any vector of remaining capacities of the resources at the beginning of a period $t$. Then on the remaining problem instance $\instance_t = \{(\reward_s, \size_s)\}_{s=t}^\numPeriod$, we consider the following estimation problem starting from time $t$ given the remaining capacity $\brconstraint$:
\begin{align}
    \label{eq:V_est_log_t}
    \tag{$\hV_{t,\brconstraint}(\instance_t)$}
    \arraycolsep=5.4pt\def\arraystretch{1.7}
    \begin{array}{llll}
    \max\limits_{\bq} ~ &
    \displaystyle
    \sum\nolimits_{j\in[\numSize]} \tsnumQuery_{j,t} \cdot \int_{1-q_j}^1 \estcdfInverse_{j,t}(u)\, \dd u 
    &
    \text{s.t.}
    \\
    &
    \displaystyle\sum\nolimits_{j\in[\numSize]} \tsnumQuery_{j,t} \cdot a_{j,i} \cdot q_j \leq \rconstraint_i
    & i\in[\numResource]
    \\
    & q_j \in [0,1] & j \in [\numSize]
    \end{array}
\end{align}
where $\estcdf_{j,t}$ is the estimator for $\cdf_j$ in time period $t$. With new input $\{\tsnumQuery_{j,t}\}_{j=1}^{\numSize}$, $\{\estcdf_{j,t}(\cdot)\}_{j=1}^{\numSize}$ and $\brconstraint$ to the Subroutine \ref{subroutine}, we can solve the optimal solution to \ref{eq:V_est_log_t} as $\{\hq_{\type,t}\}_{j=1}^{\numSize}$.
Our algorithm for poly-logarithmic regret policy (\Cref{alg:log}) is formalized as follows: 

\begin{algorithm}
\caption{\FATP}\label{alg:log}
\SetAlgoLined
\SetKwInput{KwInput}{Input}
\KwInput{
Historical samples: $\{\hat{\type}_t\}_{t=1}^T$;
Resource capacities $\constraint_i \in \reals_{\ge0}$ for every $i \in [\numResource]$;
Resource consumption vectors $\ba_j$ for every type $j \in [\numSize]$;
Reward bounds $\uppreward_j$ and $\lowreward_j$ for every type $j \in [\numSize]$;
Constant $\kappa$ defined in \Cref{th:log}.
}
Initialize remaining capacities $\constraint_{1,i} = \constraint_i$ for every $i \in [\numResource]$\;
Initialize distribution estimates $\estcdf_{j,0}$ as uniform distribution for all $j \in [\numSize]$\;

\For{$t = 1, \cdots, T$}{
Observe arrival type $\type_t$ and reward $\reward_t$\;
Update the estimation $\estcdf_{\type_t,t}$ using the new sample $\reward_t$; Set $\estcdf_{\type,t} = \estcdf_{\type,t-1}, \forall \type \neq \type_t$\; 
Compute $\tsnumQuery_{j,t}$ for each type $\type$\;
Construct the estimated relaxation problem as \ref{eq:V_est_log_t}\;
Follow Subroutine \ref{subroutine} to compute service probabilities $\hq_{\type,t}$ for every $j \in [\numSize]$\;

\If{$\hq_{\type_t,t} \geq 1-2\kappa \left(\frac{\log \numPeriod}{\sqrt{\numPeriod-t+1}}+\frac{\log \numPeriod}{\sqrt{t}}\right)$}{
    Set $M(\estcdf_{\type_t,t}, q_{\type_t,t}^{\pi}) = \lowreward_{\type_t}$\;
}
\ElseIf{$\hq_{\type_t,t} \leq 2\kappa \left(\frac{\log \numPeriod}{\sqrt{\numPeriod-t+1}}+\frac{\log \numPeriod}{\sqrt{t}}\right)$}{
    Set $M(\estcdf_{\type_t,t}, q_{\type_t,t}^{\pi}) = \uppreward_{\type_t} + 1$\;
}
\ElseIf{$2\kappa \left(\frac{\log \numPeriod}{\sqrt{\numPeriod-t+1}}+\frac{\log \numPeriod}{\sqrt{t}}\right) \leq \hq_{\type_t,t} \leq 1-2\kappa \left(\frac{\log \numPeriod}{\sqrt{\numPeriod-t+1}}+\frac{\log \numPeriod}{\sqrt{t}}\right)$}{
    Set $M(\estcdf_{\type_t,t}, q_{\type_t,t}^{\pi}) = \estcdfInverse_{\type_t,t}(1-\hq_{\type_t,t})$\;
}
Call Subroutine \ref{alg:meta} with input $(\type_t, \reward_t, M(\estcdf_{\type_t,t}, q_{\type_t,t}^{\pi}), \size_{\type_t})$ and $\constraint_{t,i}$ for every $i \in [\numResource]$.
}
\end{algorithm}

To be specific, the policy initializes remaining resource capacities and sets all reward distribution estimates to a uniform distribution. Unlike previous methods, it does not rely on historical data to pre-estimate query arrival numbers over the whole time horizon. Instead, during online execution, after observing the query type and its reward at each time period, the policy updates the corresponding reward distribution and computes an estimate of the remaining query numbers for all types. It then constructs and solves an estimated relaxation problem using the latest remaining capacities and query number estimates to obtain the current period's optimal service probabilities.
The key distinguishing feature of this policy is its threshold determination rule. Rather than directly inverting the estimated distribution function, it applies a carefully designed rounding procedure. Based on the magnitude of the computed service probability, the policy selects a threshold from three distinct regimes. This rounding mechanism is crucial for controlling approximation error dynamically throughout the horizon and is the core technique that enables the theoretical poly-logarithmic regret guarantee established in our analysis.

The proposed algorithm is proven to achieve a regret upper bound of $O((\log\numPeriod)^3)$ as following: 
\begin{theorem} \label{th:log}
With type-only samples, under \Cref{assump:reward}, \ref{assump:d} and \ref{assump:MAP2}, the regret of {\FATP} (\Cref{alg:log}) is at most
\begin{equation*}
    \left(\frac{1}{\alpha \gamma} + \frac{5 \kappa^2 (\log\numPeriod)^2}{\alpha}\right)(2\log\numPeriod+2+2\pi) + (2 s_0 + 1)\cdot \uppreward + \uppreward \cdot m = O(\sqrt{n}\cdot (\log\numPeriod)^3 + m)
\end{equation*}
where $\numResource$ represents the number of resource constraints, $\numSize$ represents the number of arrival query types and $\kappa = \frac{4\beta \sqrt{\numSize}}{\alpha \gamma} + \frac{2\beta k_1 }{\alpha \sqrt{\gamma^3}}$, $s_0 = \max\{144\kappa^2 (\log\numPeriod)^2, \frac{4}{\gamma^2 \kappa^2 (\log\numPeriod)^2}, \frac{2}{\gamma}\}$ and $\uppreward, \alpha, \beta, \gamma, k_1$ are all constants.
\end{theorem}

This result demonstrates that through fully adaptive resolving and careful rounding, poly-logarithmic regret is attainable with only minimal prior data. It thus represents a significant improvement over previous results in the literature.

\xhdr{Key Aspects of Proof Approach:}
We now highlight key aspects of our proof for achieving poly-logarithmic regret. Attaining a poly-logarithmic regret guarantee requires using the semi-fluid relaxation as a performance benchmark. A detailed discussion is provided in \Cref{apx:log}. The following lemma demonstrates that the optimal solution to our estimation problem \ref{eq:V_est_log_t} serves as a high-quality approximation of the solution to the semi-fluid relaxation.
\begin{lemma} \label{le:q_gap}
    There exists a constant $\kappa$ such that for any $\brconstraint \geq 0$, it holds that with probability at least $1-\frac{1}{\numPeriod}$,
    \begin{equation*}
        |\tq_{\type,t}^*-\hq_{\type,t}| \leq \kappa \left(\frac{\log \numPeriod}{\sqrt{s}} + \frac{\log \numPeriod}{\sqrt{\numPeriod-s+1}}\right), \forall \type \in [\numSize]
    \end{equation*}
    for remaining problem instance $\instance_t$ where $s=\numPeriod-t+1$, $\kappa = \frac{4\beta \sqrt{\numSize}}{\alpha \gamma} + \frac{2\beta k_1 }{\alpha \sqrt{\gamma^3}}$. $\{\tq_{\type,t}^*\}_{j=1}^{\numSize}$ is the optimal solution to semi-fluid relaxation problem defined in \ref{eq:V_semi_t} and $\{\hq_{\type,t}\}_{j=1}^{\numSize}$ is the optimal solution to \ref{eq:V_est_log_t}.
\end{lemma} 

The rationale for incorporating a rounding rule when computing the decision threshold is also rooted in the analysis for \Cref{th:log}. The core of our poly-logarithmic regret guarantee lies in bounding the instantaneous regret at each time period $t$. This is achieved by constructing feasible solutions to the semi-fluid benchmark through a case analysis based on the computed quantile values. The rounding mechanism is essential because it ensures the validity of these constructed solutions even when the estimated quantile $\hq_{\type_t,t}$ is extremely close to 0 or 1, which prevents the regret bound from deteriorating. A complete proof of \Cref{th:log} is also provided in \Cref{apx:log}.

\section{Conclusion and Future Directions}
\label{sec:conclusion}

We study online multi-resource allocation under arbitrary non-stationarity with a minimal possible data requirement---leveraging only a single historical sample per period. Our key contribution is a novel type-dependent quantile-based framework that cleanly decouples distribution estimation from optimization, enabling transparent, modular algorithm design.

Theoretically, we establish three principal results. First, with reward-observed samples, our proposed {\STP} achieves $\tilde{O}(\sqrt{T})$ regret in the multi-resource setting. Second, for type-only samples, we prove that sublinear regret is impossible without structural assumptions, but becomes attainable under a mild minimum-arrival-probability condition by our proposed {\PATP}. Third and most significantly, we design {\FATP} with careful rounding that achieves the \emph{first poly-logarithmic regret guarantee} of ${O}((\log T)^3)$ for non-stationary multi-resource allocation---a qualitative improvement over all prior dual-based approaches.

Beyond regret bounds, our quantile-based paradigm offers conceptual advantages: decisions are type-specific with no cross-type interference, the framework is modular (allowing plug-and-play estimation methods), and the policy logic is transparent (``accept rewards above the $(1-q_j)$-th quantile''). This work demonstrates that near-optimal performance in volatile environments requires only the minimal possible offline information---a single sample per period suffices for remarkably strong guarantees.

Future directions include extending our framework to settings with reusable resources, contextual information, or bandit feedback where rewards are only observed upon acceptance.
Finally, empirical validation on real-world non-stationary workloads would further substantiate the practical relevance of our theoretically grounded approach.

\bibliographystyle{abbrvnat}
\bibliography{refs}

\clearpage

\OneAndAHalfSpacedXI

\begin{APPENDICES}

\section{Further Related Work}
\label{apx:related work}
We discuss further related work to ours here:

\xhdr{Online Resource Allocation:}
The online resource allocation problem, commonly referred to as the AdWords problem, has been extensively studied. Foundational work by \citet{MSVV-07} introduced the trade-off revealing linear program and derived an optimal algorithm with a competitive ratio of $1-1/e$. \citet{BJN-07} developed a primal-dual framework that achieved the same optimal ratio. This line of research was significantly generalized by \citet{DJ-12}, who allowed for arbitrary concave returns and characterized the optimal competitive ratio. \citet{KP-16} further extended the primal-dual analysis, demonstrating that a constant competitive ratio is impossible in the most general setting and providing asymptotically tight bounds.
Beyond competitive analysis, structural and algorithmic insights have been developed for related dynamic scheduling problems. For instance, \citet{FLTZ-14} proposed heuristic procedures for online advance scheduling, while \citet{Tru-15} provided analytical results for a two-class model. A notable algorithmic framework was introduced by \citet{VBG-21}, which designed simple yet efficient policies for a broad family of problems including online packing, budget-constrained probing, and contextual bandits with knapsacks. From a learning-theoretic perspective, \citet{BLM-23} employed dual mirror descent to achieve sublinear regret relative to the hindsight optimum for online allocation with concave rewards.

To handle uncertainty in reward or consumption distributions, online learning techniques become essential. \citet{PSZT-20} studied online matching with unknown reward distributions, proposing a two-phase explore-then-exploit algorithm for non-stationary Poisson arrivals. Similarly, \citet{CMSW-22} combined inventory balancing with online learning to allocate resources to heterogeneous customers despite unknown consumption patterns.
For non-stationary environments, \citet{BCKZ-19} proposed a time-segmentation approach that converts a non-stationary problem into a sequence of stationary sub-problems. More generally, \citet{ZKSV-22} introduced a framework that jointly optimizes exploration, exploitation, and robustness against adversarial arrival sequences, offering a principled approach to dynamic resource allocation under distributional shifts.

\xhdr{Non-stationary Stochastic Optimization:}
A growing body of literature addresses sequential decision-making under non-stationarity. Early work in network revenue management (NRM) tackled time-varying arrivals through approximate dynamic programming; for instance, \citet{Ade-07} derived a deterministic linear program for bid-price control using such an approach. Subsequent refinements, such as the compact linear programming formulation for piecewise-linear value function approximations by \citet{KT-16}, offered improved computational tractability.
A prominent framework for quantifying and managing non-stationarity is the variation budget, introduced by \citet{BGZ-15} for sequential stochastic optimization with changing cost functions. This concept was later generalized by \citet{CWW-19}, who derived matching upper and lower regret bounds for smooth, strongly convex function sequences under $L_{p,q}$-variation constraints.
In parallel, research on non-stationary Markov Decision Processes (MDPs) has extended these ideas to reinforcement learning. \citet{LR-19} studied model-based algorithms in evolving MDPs, while \citet{CSZ-23} developed a sliding-window upper confidence bound algorithm with a confidence-widening technique, establishing dynamic regret bounds when variation budgets are known.

\xhdr{MAB with Resource Constraints:}
Our work is further connected to the literature on multi-armed bandits (MAB) under resource constraints, commonly known as bandits with knapsacks (BwK). This framework generalizes the classical MAB problem by incorporating long-term resource consumption constraints alongside reward maximization. The BwK model was formally introduced by \citet{BKS-18}.
Subsequent research has significantly expanded its scope and developed efficient algorithms. \citet{AD-14} considered a general model accommodating concave rewards and convex constraints. The contextual setting, where side information is available per round, has been extensively studied: \citet{BLS-14} examined contextual MAB under budget constraints, while \citet{ADL-16} and \citet{AD-16} developed efficient algorithms employing confidence ellipsoids for parameter estimation. From a regret analysis perspective, \citet{LSY-21} designed a primal-dual algorithm achieving a problem-dependent logarithmic regret bound. For adversarial environments, \citet{ISSS-22} derived an algorithm with an $O(\log T)$ competitive ratio relative to the best fixed action distribution, a result later improved upon by \citet{CCK-22}.
The challenge of non-stationarity within this constrained bandit setting has also been addressed. \citet{BGZ-14} proposed a tractable MAB formulation allowing reward distributions to change over time. More directly related to our setting, \citet{LJL-22} study BwK in a non-stationary environment, providing a primal-dual analysis that explicitly characterizes the interplay between resource constraints and distribution shifts.

\section{Estimation of Arrival Distributions}
\label{apx:est_arrival-distribution}
\xhdr{Proof of \Cref{le:est_dj}:}

For every $t$, we obtain one single sample $\stype_t \sim \typedf_t$. Since the query types are finite, i.e. $\stype_t \in [\numSize]$, we define the number of type $\type$ query arrivals from historical observation as $\snumQuery_j = \sum_{t=1}^\numPeriod \indicator{\stype_t=j}$. Clearly, $\sum_{\type=1}^\numSize \snumQuery_j = \numPeriod$ as we have exactly $\numPeriod$ samples in total. For a given problem instance $\instance$, the actual number of type $\type$ queries is $\numQuery_j = \sum_{t=1}^\numPeriod \indicator{\type_t=j}$. Since $\type_t$ and $\stype_t$ are independent and identically distributed according to $\typedf_t$, it follows that:
\begin{equation*}
    \expect[\instance \sim G]{\indicator{\type_t=j}} = \typedf_t(j), \quad \expect[\instance \sim G]{\indicator{\stype_t=j}} = \typedf_t(j).
\end{equation*}

Consequently, for any $\type \in [\numSize]$ we have
\begin{equation*}
    \expect[\instance \sim G]{\numQuery_j} = \sum_{t=1}^\numPeriod \typedf_t(j) = \expect[\instance \sim G]{\snumQuery_j}.
\end{equation*}
Therefore $\snumQuery_j$ is an unbiased estimator for $\expect[\instance \sim G]{\numQuery_j}$. For simplification, we define $\mu_j = \expect[\instance \sim G]{\numQuery_j}$ in the following analysis.
Applying Hoeffding Inequality, for any $\epsilon > 0$, we have
\begin{equation*}
    \prob{|\snumQuery_j - \mu_j| \geq \epsilon} = \prob{|\snumQuery_j - \expect[\instance \sim G]{\snumQuery_j}| \geq \epsilon}  \leq 2 \exp\left( - \frac{2 \epsilon^2}{\sum_{t=1}^\numPeriod (1 - 0)^2} \right) = 2 \exp\left( - \frac{2 \epsilon^2}{\numPeriod} \right).
\end{equation*}

Thus, with probability at least $1 - \delta$,
\begin{equation*}
    |\snumQuery_j - \mu_j| \leq \sqrt{\frac{\numPeriod}{2} \log(2/\delta)}.
\end{equation*}

By Cauchy-Schwarz Inequality, we have
\begin{equation*}
    \expect[\instance \sim G]{\sum_{\type=1}^\numSize |\snumQuery_j - \mu_j|} \leq \sum_{\type=1}^\numSize \sqrt{\Var(\snumQuery_j)} \leq \sqrt{\numSize \sum_{\type=1}^\numSize \Var(\snumQuery_j)} \leq \sqrt{\numSize \numPeriod}
\end{equation*}
where the last inequality holds for $\sum_{\type=1}^\numSize \Var(\snumQuery_j) = \sum_{\type=1}^\numSize \sum_{t=1}^\numPeriod \typedf_t(j)(1 - \typedf_t(j)) \leq \sum_{\type=1}^\numSize \sum_{t=1}^\numPeriod \typedf_t(j) = \numPeriod$.

Applying McDiarmid Inequality, for any $\epsilon > 0$, we have
\begin{equation*}
    \prob{|\sum_{\type=1}^\numSize |\snumQuery_j - \mu_j| - \expect[\instance \sim G]{\sum_{\type=1}^\numSize |\snumQuery_j - \mu_j|}| \geq \epsilon}  \leq 2 \exp\left( - \frac{2\epsilon^2}{4\numPeriod} \right) = 2 \exp\left( - \frac{\epsilon^2}{2\numPeriod} \right).
\end{equation*}

With probability at least $1 - \delta$, we have
\begin{equation} \label{eq:est_dj_1}
    \sum_{\type=1}^\numSize |\snumQuery_j - \mu_j| \leq \sqrt{\numSize \numPeriod} + \sqrt{2\numPeriod \log(2/\delta)} \leq \sqrt{2\numSize \numPeriod \log(2/\delta)}
\end{equation}

We now consider the upper bound for $\sum_{\type=1}^\numSize |\snumQuery_j - \mu_j|^2/\mu_j$. 
By Bernstein Inequality, for any $\epsilon > 0$, we have
\begin{equation*}
    \prob{|\snumQuery_j - \mu_j| \geq \epsilon} \leq 2 \exp\left( - \frac{\epsilon^2}{2(\Var(\snumQuery_j)+\epsilon/3)} \right)
\end{equation*}

Let $\epsilon = L_1 \sqrt{\mu_j}$ where $L_1$ is a constant to be determined, we have
\begin{equation*}
    \prob{|\snumQuery_j - \mu_j| \geq L_1 \sqrt{\mu_j}} = \prob{|\snumQuery_j - \mu_j|^2/\mu_j \geq L_1^2} \leq 2 \exp\left( - \frac{L_1^2 \mu_j}{2(\mu_j + L_1 \sqrt{\mu_j}/3)} \right)
\end{equation*}

From \Cref{assump:d}, for any $j \in [\numSize]$, $\prob{\snumQuery_j \geq 1} = 1$, we know that $\mu_j \geq 1$ always holds. Therefore
\begin{equation*}
    \prob{\frac{|\snumQuery_j - \mu_j|^2}{\mu_j} \geq L_1^2} \leq 2 \exp\left( - \frac{L_1^2 \mu_j}{2(\mu_j + L_1 \mu_j/3)} \right) = 2 \exp\left( - \frac{L_1^2}{2(1 + L_1/3)} \right)
\end{equation*}

Consequently, with probability at least $1 - \delta$, we know that 
\begin{equation*} 
    \frac{|\snumQuery_j - \mu_j|^2}{\mu_j} \leq \left(\frac{1}{3}\log(\frac{2}{\delta})+ \sqrt{\frac{1}{9}(\log(\frac{2}{\delta}))^2 + 2\log(\frac{2}{\delta})}\right)^2 \leq \left(\frac{2}{3}\log(\frac{2}{\delta})+3\right)^2
\end{equation*}

Therefore, for $\delta < 1/2$, with probability at least $1 - \delta$, we have
\begin{equation}\label{eq:est_dj_2}
    \sum_{\type=1}^\numSize \frac{|\snumQuery_j - \mu_j|^2}{\mu_j} \leq \numSize \left(\frac{2}{3}\log(\frac{2}{\delta})+3\right)^2 \leq \numSize \left(4\log(\frac{2}{\delta})\right)^2
\end{equation}

\xhdr{Proof of \Cref{le:est_Fj}:}

We define the kernel estimation as:
\begin{equation*}
    \estcdf(x) = \frac{1}{\sample} \sum_{i=1}^{\sample} K(\frac{x - X_i}{h_{\sample}})
\end{equation*}
where $h_{\sample} > 0$ is the bandwidth parameter to be specified. We split $\estcdf(x) - \cdf(x)$ into two parts: the bias $\expect{\estcdf(x)} - \cdf(x)$ and the random fluctuation $\estcdf(x) - \expect{\estcdf(x)}$.

We first discuss the upper bound for bias $\expect{\estcdf(x)} - \cdf(x)$.
Take expectation of $\estcdf(x)$ with respect to the randomness in the samples $\{X_1, \cdots, X_{\sample}\}$ which are independent and identically distributed from the true distribution function $\cdf(x)$, we have
\begin{equation*}
    \begin{aligned}
        \expect{\estcdf(x)} & = \expect[X_1, \cdots, X_{\sample}]{\frac{1}{\sample} \sum_{i=1}^{\sample} K(\frac{x - X_i}{h_{\sample}})}\\
        & = \expect[X_1]{K(\frac{x - X_1}{h_{\sample}})} \\
        & = \int_{-\infty}^{+\infty} K(\frac{x - y}{h_{\sample}}) \pdf(y)\, \dd y
    \end{aligned}
\end{equation*}
where $\pdf$ is the probability density function of $\cdf$.
Define $u=(x-y)/h_{\sample}$, we have
\begin{equation*}
    \expect{\estcdf(x)} = \int_{-\infty}^{+\infty} K(\frac{x - y}{h_{\sample}}) \pdf(y)\, \dd y = h_{\sample} \int_{-\infty}^{+\infty} K(u) \pdf(x- h_{\sample}u)\, \dd u
\end{equation*}

Meanwhile,
\begin{equation*}
    \cdf(x) = \int_{-\infty}^{+\infty} \indicator{y \leq x} \pdf(y)\, \dd y = h_{\sample} \int_{-\infty}^{+\infty} \indicator{u \geq 0} \pdf(x- h_{\sample}u)\, \dd u
\end{equation*}

We define $L(u) = K(u) - \indicator{u \geq 0}$. Since $k(u)$ is supported on [-1,1], we know $K(u)=0$ when $u \leq -1$, and $K(u)=1$ when $u \geq 1$. Consequently, $L(u)$ is also supported on [-1,1]. Therefore,
\begin{equation*}
    \begin{aligned}
        \expect{\estcdf(x)} - \cdf(x) & = h_{\sample} \int_{-\infty}^{+\infty} L(u) \pdf(x- h_{\sample}u)\, \dd u\\
        & = h_{\sample} \int_{-1}^{1} L(u) \pdf(x- h_{\sample}u)\, \dd u
    \end{aligned}
\end{equation*}

We prove that $\int_{-1}^{1} |L(u)|\, \dd u$ is bounded: 
Since $K(u)$ is symmetric, we know $K(-u) = 1-K(u)$ and $K(0)=\frac{1}{2}$.
Since $K(u)$ is monotone increasing, we know $0 \leq K(u) \leq \frac{1}{2}$ when $-1 \leq u \leq 0$, and $\frac{1}{2} \leq K(u) \leq 1$ when $0 \leq u \leq 1$. Therefore,
\begin{equation*}
    \begin{aligned}
        \int_{-1}^{1} |L(u)|\, \dd u
        & = \int_{-1}^{0} K(u)\, \dd u + \int_{0}^{1} [1-K(u)]\, \dd u\\
        & \leq \int_{-1}^{0} \frac{1}{2}\, \dd u + \int_{0}^{1} \frac{1}{2}\, \dd u\\
        & = 1
    \end{aligned} 
\end{equation*}

\textbf{Case 1: Interior points.} When $x \in [a+h_{\sample},b-h_{\sample}]$, for all $u \in [-1,1]$, we have $x- h_{\sample}u \in [a,b]$, so $\pdf(x- h_{\sample}u)$ is well-defined and upper bounded by constant $\beta$. Hence,
\begin{equation*}
    \expect{\estcdf(x)} - \cdf(x) \leq h_{\sample} \int_{-1}^{1} |L(u)| |\pdf(x- h_{\sample}u)|\, \dd u \leq \beta h_{\sample} \int_{-1}^{1} |L(u)|\, \dd u \leq \beta h_{\sample} 
\end{equation*}

\textbf{Case 2: Boundary regions.} When $x \in [a,a+h_{\sample}) \cup (b-h_{\sample},b]$, we first consider the left boundary $x=a+\theta h_{\sample}$ where $0\leq \theta<1$. Therefore, when $u>\theta$, we have $x- h_{\sample}u <a$ and $\pdf(x- h_{\sample}u)=0$. Split the integral into two parts:
\begin{equation*}
    \begin{aligned}
        \expect{\estcdf(x)} - \cdf(x) & = h_{\sample} \int_{-1}^{\theta} L(u) \pdf(x- h_{\sample}u)\, \dd u + h_{\sample} \int_{\theta}^{1} L(u) \pdf(x- h_{\sample}u)\, \dd u \\
        & \leq h_{\sample} \int_{-1}^{\theta} |L(u)| |\pdf(x- h_{\sample}u)|\, \dd u\\
        & \leq \beta h_{\sample}
    \end{aligned}
\end{equation*}

Similarly, the same conclusion holds for the right boundary $x \in (b-h_{\sample},b]$.
To conclude, 
\begin{equation} \label{eq:est_Fj_1}
    \sup_{x \in [a,b]} |\expect{\estcdf(x)} - \cdf(x)| \leq \beta h_{\sample}
\end{equation}

Now we consider the upper bound for random fluctuation $\estcdf(x) - \expect{\estcdf(x)}$. We define the empirical distribution function as $\empcdf(x)$. We have
\begin{equation*}
    \estcdf(x) = \frac{1}{\sample} \sum_{i=1}^{\sample} \int_{-\infty}^{\frac{x - X_i}{h_{\sample}}} k(u)\, \dd u = \int_{-\infty}^{+\infty} \empcdf(x- h_{\sample}u) k(u)\, \dd u
\end{equation*}

Similarly,
\begin{equation*}
    \expect{\estcdf(x)} = \int_{-\infty}^{+\infty} \cdf(x- h_{\sample}u) k(u)\, \dd u
\end{equation*}

Therefore,
\begin{equation*}
    \begin{aligned}
        |\estcdf(x) - \expect{\estcdf(x)}| & \leq \int_{-\infty}^{+\infty} |\empcdf(x- h_{\sample}u) - \cdf(x- h_{\sample}u)| k(u)\, \dd u\\
        & \leq \sup_{y \in [a,b]}|\empcdf(y) - \cdf(y)| \int_{-\infty}^{+\infty} k(u)\, \dd u\\
        & = \sup_{y \in [a,b]}|\empcdf(y) - \cdf(y)|
    \end{aligned}
\end{equation*}

Consequently,
\begin{equation*}
    \sup_{x \in [a,b]}|\estcdf(x) - \expect{\estcdf(x)}| \leq \sup_{y \in [a,b]}|\empcdf(y) - \cdf(y)|
\end{equation*}

Applying Dvoretzky–Kiefer–Wolfowitz Inequality, for any $\epsilon>0$, we have
\begin{equation*}
    \prob{\sup_{y \in [a,b]}|\empcdf(y) - \cdf(y)| \geq \epsilon} \leq 2 \exp\left( - 2 \sample \epsilon^2 \right) 
\end{equation*}

Thus, with probability at least $1-\delta$,
\begin{equation} \label{eq:est_Fj_2}
    \sup_{x \in [a,b]}|\estcdf(x) - \expect{\estcdf(x)}| \leq \sup_{y \in [a,b]}|\empcdf(y) - \cdf(y)| \leq \sqrt{\frac{\log(2/\delta)}{2\sample}}
\end{equation}

Combining (\ref{eq:est_Fj_1}) and (\ref{eq:est_Fj_2}), we have that with probability at least $1-\delta$,
\begin{equation*}
    \begin{aligned}
        \sup_{x \in [a,b]} |\cdf(x) - \estcdf(x)| & \leq \sup_{x \in [a,b]} |\expect{\estcdf(x)} - \cdf(x)| + \sup_{x \in [a,b]}|\estcdf(x) - \expect{\estcdf(x)}|\\
        & \leq \beta h_{\sample} + \sqrt{\frac{\log(2/\delta)}{2\sample}}
    \end{aligned}
\end{equation*}

We set $h_{\sample} = \sample^{-1/2}$, then with probability at least $1-\delta$,
\begin{equation*}
    \sup_{x \in [a,b]} |\cdf(x) - \estcdf(x)| \leq (\beta + 1)\sqrt{\frac{\log(1/\delta)}{\sample}} = O(\sqrt{\frac{\log(1/\delta)}{\sample}})
\end{equation*}

\section{Missing Proof in \texorpdfstring{\Cref{sec:offline setting}}{Section 4}}
\label{apx:offline}
\xhdr{Proof of \Cref{th:offline}:}

We rewrite the \emph{regret} according to \ref{eq:V_fluid2}:
\begin{equation*}
    \begin{aligned}
        \regret(\pi) & \leq \bVfluid_{\bconstraint} - \expect[\instance \sim G]{V_{\bconstraint}^{\pi}(\instance)}\\
        & = \expect[\instance \sim G]{\sum_{j=1}^\numSize \numQuery_j (\int_{1-q_j^*(\blambda^*)}^1 \cdfInverse_j(u)\, \dd u - \int_{1-q_j^{\pi}}^1 \cdfInverse_j(u)\, \dd u)}\\
        & \leq \expect[\instance \sim G]{\sum_{j=1}^\numSize \mu_j \cdot \uppreward_j \cdot |q_j^*(\blambda^*) - q_j^{\pi}|}
    \end{aligned}
\end{equation*}
where $q_j^{\pi}$ denotes the probability that query $\query$ will be served by the online policy $\pi$ and the inequality holds because the reward for type $\type$ is upper bounded by $\uppreward_j$. The most straightforward approach to implementing $q_j^{\pi}$ is to set $q_j^{\pi} := \hq_j(\hblambda)$. However, this would result in constraint violations in practice. Therefore, we reformulate the regret of our algorithm as follows:
\begin{equation}\label{eq:offline_regret}
    \begin{aligned}
        \regret(\pi) & \leq \expect[\instance \sim G]{\sum_{j=1}^\numSize \mu_j \cdot \uppreward_j \cdot |q_j^*(\blambda^*) - q_j^{\pi}|}\\
        & \leq \expect[\instance \sim G]{\sum_{j=1}^\numSize \mu_j \cdot \uppreward_j \cdot |q_j^*(\blambda^*) - \hq_j(\hblambda)| +  V(\Hat{\bq}(\hblambda))}
    \end{aligned}
\end{equation}
where $V(\Hat{\bq}(\hblambda))$ denotes the penalty incurred due to constraint violations when directly applying $\{\hq_j(\hblambda)\}_{j=1}^\numSize$ in our algorithm.
For $V(\Hat{\bq}(\hblambda))$, we have the following upper bound:
\begin{equation}\label{eq:Violation_offline}
   \begin{aligned}
       V(\Hat{\bq}(\hblambda)) & \leq \uppreward \sum_{i=1}^\numResource \max \{0, \sum_{j=1}^\numSize \mu_j a_{j,i} \hq_j(\hblambda) - \constraint_i\}\\
       & \leq \uppreward \sum_{i=1}^\numResource \max \{0, \sum_{j=1}^\numSize \mu_j a_{j,i} \hq_j(\hblambda) - \sum_{j=1}^\numSize \mu_j a_{j,i} q_j^*(\blambda^*)\}\\
       & \leq \uppreward \sum_{i=1}^\numResource \sum_{j=1}^\numSize \mu_j a_{j,i} |q_j^*(\blambda^*) - \hq_j(\hblambda)|\\
       & \leq \sum_{j=1}^\numSize \mu_j \cdot \uppreward \numResource \amax |q_j^*(\blambda^*) - \hq_j(\hblambda)|
   \end{aligned} 
\end{equation}
where $\uppreward = \max_j \uppreward_j, \amax = \max_{j,i} a_{j,i}$ and the second inequality follows from the budget constraints.

We now proceed to bound the term $\sum_{j=1}^\numSize \mu_j |q_j^*(\blambda^*) - \hq_j(\hblambda)|$.
Recall from (\ref{eq:q_optimal}) that $\{q_j^*(\blambda^*)\}_{j=1}^\numSize$ is the optimal solution to \ref{eq:V_fluid2} and (\ref{eq:q_offline}) that $\{\hq_j(\hblambda)\}_{j=1}^\numSize$ is the optimal solution to \ref{eq:V_est_offline}. For notational simplicity, we will omit the dependence on $\blambda$ and write $q_j^*(\blambda^*)$ as $q_j^*$, $\hq_j(\hblambda)$ as $\hq_j$ n the subsequent analysis.

For each $j \in [\numSize]$, we define
\begin{equation*}
    G_j(q) = \int_{1-q}^1 \cdfInverse_j(u)\, \dd u
\end{equation*}

Since we assume the lower bound and upper bound of pdf $\pdf_j$, we have
\begin{equation} \label{eq:Gj_prop}
    G_j'(q) = \cdfInverse_j(1-q) , \ \ G_j''(q) = - \frac{1}{\pdf_j(\cdfInverse_j(1-q))} \leq - \frac{1}{\beta}
\end{equation}
thus $G_j(q)$ is $1/\beta$-strongly concave and $G_j'(q)$ is $1/\alpha$-Lipschitz-continuous.

To address the regret arising from estimation errors in both query numbers and reward distributions, we introduce an intermediate variable $\tq_j$ and decompose $|q_j^* - \hq_j|$ into two parts using the triangle inequality.

Specifically, we consider an intermediate estimation problem in which the reward distributions are known accurately:
\begin{align}
    \label{eq:V_inter_offline}
    \tag{$\tV_{\bconstraint}(\bsnumQuery)$}
    \arraycolsep=5.4pt\def\arraystretch{1.7}
    \begin{array}{llll}
    \max\limits_{\bq} ~ &
    \displaystyle
    \sum\nolimits_{j\in[\numSize]} \snumQuery_j \cdot \int_{1-q_j}^1 \cdfInverse_j(u)\, \dd u 
    &
    \text{s.t.}
    \\
    &
    \displaystyle\sum\nolimits_{j\in[\numSize]} \snumQuery_j \cdot a_{j,i} \cdot q_j \leq \constraint_i
    & i\in[\numResource]
    \\
    & q_j \in [0,1] & j \in [\numSize]
    \end{array}
\end{align}
and let $\{\tq_j\}_{j=1}^\numSize$ denote its optimal solution. Then we have 
\begin{equation} \label{eq:inter_offline}
    \sum_{\type=1}^\numSize \mu_j |q_j^* - \hq_j| \leq \sum_{\type=1}^\numSize \mu_j |q_j^* - \tq_j| + \sum_{\type=1}^\numSize \mu_j |\tq_j - \hq_j|
\end{equation}

For the first term on the right-hand side of (\ref{eq:inter_offline}), we use the strong concavity of $G_j$, which yields
\begin{equation*}
    \frac{1}{\beta}(\tq_j - q_j^*)^2 \leq  \left(G_j'(\tq_j)-G_j'(q_j^*)\right)(q_j^* - \tq_j)
\end{equation*}

We define the optimal dual variable to \ref{eq:V_fluid2} as $\blambda^*$ and the optimal dual variable to \ref{eq:V_inter_offline} as $\tblambda$. From first-order condition, we have
\begin{equation*}
    G_j'(q_j^*) = \sum_{i=1}^{\numResource} \lambda_i^* a_{j,i}, \quad G_j'(\tq_j) = \sum_{i=1}^{\numResource} \tlambda_i a_{j,i}
\end{equation*}

From this it follows that
\begin{equation*}
    \begin{aligned}
        \frac{1}{\beta} \sum_{\type=1}^\numSize \mu_j (\tq_j - q_j^*)^2 & \leq \sum_{\type=1}^\numSize \mu_j \left(G_j'(\tq_j)-G_j'(q_j^*)\right)(q_j^* - \tq_j)\\
        & = \sum_{\type=1}^\numSize \mu_j \sum_{i=1}^{\numResource} (\tlambda_i - \lambda_i^*)a_{j,i}(q_j^* - \tq_j)\\
        & = \sum_{i=1}^{\numResource} (\tlambda_i - \lambda_i^*) \left(\sum_{\type=1}^\numSize \mu_j a_{j,i}(q_j^* - \tq_j)\right)
    \end{aligned}
\end{equation*}

Define $\Delta\constraint_i = \sum_{\type=1}^\numSize \mu_j a_{j,i}q_j^* - \sum_{\type=1}^\numSize \snumQuery_j a_{j,i}\tq_j$ for every $i \in [\numResource]$, we have
\begin{equation*}
    \sum_{i=1}^{\numResource} (\tlambda_i - \lambda_i^*) \left(\sum_{\type=1}^\numSize \mu_j a_{j,i}(q_j^* - \tq_j)\right) = \sum_{i=1}^{\numResource} (\tlambda_i - \lambda_i^*) \cdot \Delta\constraint_i + \sum_{i=1}^{\numResource} (\tlambda_i - \lambda_i^*) \left(\sum_{\type=1}^\numSize (\mu_j-\snumQuery_j) a_{j,i}\tq_j\right)
\end{equation*}

From complementary slackness and resource constraints, we have
\begin{equation*}
    \begin{aligned}
        & \lambda_i^* (\sum_{\type=1}^\numSize \mu_j a_{j,i}q_j^* - \constraint_i) = 0, \quad \sum_{\type=1}^\numSize \mu_j a_{j,i}q_j^* \leq \constraint_i\\
        & \tlambda_i (\sum_{\type=1}^\numSize \snumQuery_j a_{j,i}\tq_j - \constraint_i) = 0, \quad \sum_{\type=1}^\numSize \snumQuery_j a_{j,i}\tq_j \leq \constraint_i
    \end{aligned}
\end{equation*}
when $\lambda_i^* > 0$, it holds that $\sum_{\type=1}^\numSize \mu_j a_{j,i}q_j^* = \constraint_i$, thus $\Delta\constraint_i \geq 0$; when $\tlambda_i > 0$, it holds that $\sum_{\type=1}^\numSize \snumQuery_j a_{j,i}\tq_j \leq \constraint_i$, thus $\Delta\constraint_i \leq 0$. Consequently we have
\begin{equation*}
    \sum_{i=1}^{\numResource} (\tlambda_i - \lambda_i^*) \cdot \Delta\constraint_i \leq 0
\end{equation*}

Therefore, we have
\begin{equation*}
    \begin{aligned}
        \frac{1}{\beta} \sum_{\type=1}^\numSize \mu_j (\tq_j - q_j^*)^2 & \leq  \sum_{i=1}^{\numResource} (\tlambda_i - \lambda_i^*) \cdot \Delta\constraint_i + \sum_{i=1}^{\numResource} (\tlambda_i - \lambda_i^*) \left(\sum_{\type=1}^\numSize (\mu_j-\snumQuery_j) a_{j,i}\tq_j\right)\\
        & \leq \sum_{\type=1}^\numSize (\mu_j-\snumQuery_j) \tq_j \sum_{i=1}^{\numResource} (\tlambda_i - \lambda_i^*)a_{j,i}\\
        & = \sum_{\type=1}^\numSize (\mu_j-\snumQuery_j) \tq_j \left(G_j'(\tq_j)-G_j'(q_j^*)\right)\\
        & \leq \sum_{\type=1}^\numSize |\mu_j-\snumQuery_j| |G_j'(\tq_j)-G_j'(q_j^*)|
    \end{aligned}
\end{equation*}

Since $G_j'(q)$ is $1/\alpha$-Lipschitz-continuous, we know that
\begin{equation*}
    |G_j'(\tq_j)-G_j'(q_j^*)| \leq \frac{1}{\alpha} |\tq_j - q_j^*|
\end{equation*}

Thus
\begin{equation*}
    \frac{1}{\beta} \sum_{\type=1}^\numSize \mu_j (\tq_j - q_j^*)^2 \leq \frac{1}{\alpha} \sum_{\type=1}^\numSize |\mu_j - \snumQuery_j||\tq_j - q_j^*|
\end{equation*}

By Cauchy-Schwarz Inequality,  we have
\begin{equation*}
    \begin{aligned}
        \frac{1}{\beta} \sum_{\type=1}^\numSize \mu_j (\tq_j - q_j^*)^2 & \leq \frac{1}{\alpha} \sum_{\type=1}^\numSize |\snumQuery_j - \mu_j||\tq_j - q_j^*|\\
        & \leq \frac{1}{\alpha} (\sum_{\type=1}^\numSize \mu_j (\tq_j - q_j^*)^2)^{\frac{1}{2}} (\sum_{\type=1}^\numSize \frac{|\snumQuery_j - \mu_j|^2}{\mu_j})^{\frac{1}{2}}
    \end{aligned} 
\end{equation*}

Therefore,
\begin{equation*} 
    \sum_{\type=1}^\numSize \mu_j (\tq_j - q_j^*)^2 \leq (\frac{\beta}{\alpha})^2 \sum_{\type=1}^\numSize \frac{|\snumQuery_j - \mu_j|^2}{\mu_j}
\end{equation*}

From \Cref{apx:est_arrival-distribution}, we have (\ref{eq:est_dj_2}). Following Cauchy-Schwarz Inequality, with probability at least $1-\frac{1}{\numPeriod}$, we have
\begin{equation}\label{eq:upper_bound_term1}
    \begin{aligned}
        \sum_{\type=1}^\numSize \mu_j |\tq_j - q_j^*| & \leq (\sum_{\type=1}^\numSize \mu_j )^{\frac{1}{2}} (\sum_{\type=1}^\numSize \mu_j (\tq_j - q_j^*)^2)^{\frac{1}{2}}\\
        & \leq \sqrt{\numPeriod} \cdot \frac{\beta}{\alpha} (\sum_{\type=1}^\numSize \frac{|\snumQuery_j - \mu_j|^2}{\mu_j})^{\frac{1}{2}}\\
        & \leq \sqrt{\numPeriod} \cdot \frac{\beta}{\alpha} \cdot 4\log(2\numPeriod) \sqrt{\numSize}\\
        & \leq \frac{4\beta}{\alpha} \log(\numPeriod) \sqrt{\numSize \numPeriod}
    \end{aligned} 
\end{equation}

For the second term on the right-hand side of (\ref{eq:inter_offline}), we define
\begin{equation*}
    \hG_j (q) = \int_{1-q}^1 \estcdfInverse_j(u)\, \dd u
\end{equation*}

We can also apply the strongly concavity of $G_j$:
\begin{equation*}
    G_j (\tq_j) \leq G_j(\hq_j) + G_j'(\hq_j)(\tq_j - \hq_j) - \frac{1}{2\beta}(\tq_j - \hq_j)^2
\end{equation*}

Thus
\begin{equation} \label{eq:optimal_cdf_gap_bound}
    \begin{aligned}
        \frac{1}{2\beta} \sum_{\type=1}^\numSize \snumQuery_j (\tq_j - \hq_j)^2 & \leq \sum_{\type=1}^\numSize \snumQuery_j G_j(\hq_j) - \sum_{\type=1}^\numSize \snumQuery_j G_j(\tq_j) + \sum_{\type=1}^\numSize \snumQuery_j G_j'(\hq_j)(\tq_j - \hq_j)\\
        & \leq \sum_{\type=1}^\numSize \snumQuery_j G_j'(\hq_j)(\tq_j - \hq_j)\\
        & = \sum_{\type=1}^\numSize \snumQuery_j \hG_j'(\hq_j)(\tq_j - \hq_j) + \sum_{\type=1}^\numSize \snumQuery_j (G_j'(\hq_j)-\hG_j'(\hq_j))(\tq_j - \hq_j)\\
        & \leq \sum_{\type=1}^\numSize \snumQuery_j |\cdfInverse_j(1-\hq_j) - \estcdfInverse_j(1-\hq_j)||\tq_j - \hq_j|\\
        & \leq \sum_{\type=1}^\numSize \snumQuery_j \frac{\epsilon_j}{\alpha} |\tq_j - \hq_j|
    \end{aligned}
\end{equation}
where  $\epsilon_j = \sup_x |\estcdf_j(x) - \cdf_j(x)|$, the second and third inequality hold for the optimality of $\{\tq_j\}_{j=1}^\numSize$ to \ref{eq:V_inter_offline} and $\{\hq_j\}_{j=1}^\numSize$ to \ref{eq:V_est_offline}.
From \Cref{le:est_Fj}, with probability at least $1-\frac{1}{\numPeriod}$, we have that $\epsilon_j = \sup_x |\estcdf_j(x) - \cdf_j(x)| \leq k_1 \sqrt{\log(\numPeriod) / \snumQuery_j} = O(\sqrt{\log (\numPeriod) / \snumQuery_j})$ where $k_1$ is a constant. Therefore by Cauchy-Schwarz Inequality, we have
\begin{equation*}
    \begin{aligned}
        \frac{1}{2\beta} \sum_{\type=1}^\numSize \snumQuery_j (\tq_j - \hq_j)^2 & \leq \sum_{\type=1}^\numSize \snumQuery_j \frac{\epsilon_j}{\alpha} |\tq_j - \hq_j|\\
        & \leq \frac{1}{\alpha} (\sum_{\type=1}^\numSize \snumQuery_j (\tq_j - \hq_j)^2)^{\frac{1}{2}} (\sum_{\type=1}^\numSize \snumQuery_j (\epsilon_j^2))^{\frac{1}{2}}\\
        & \leq \frac{k_1 \sqrt{\numSize \log(\numPeriod)}}{\alpha} \sum_{\type=1}^\numSize \snumQuery_j (\tq_j - \hq_j)^2)^{\frac{1}{2}}
    \end{aligned} 
\end{equation*}

Thus with probability at least $1-\frac{1}{\numPeriod}$,
\begin{equation*}
    \sum_{\type=1}^\numSize \snumQuery_j (\tq_j - \hq_j)^2 \leq (\frac{2\beta k_1}{\alpha})^2 \cdot \numSize \log(\numPeriod)
\end{equation*}

By Cauchy-Schwarz Inequality,  we have
\begin{equation*}
    \begin{aligned}
        \sum_{\type=1}^\numSize \snumQuery_j |\tq_j - \hq_j| & \leq (\sum_{\type=1}^\numSize \snumQuery_j )^{\frac{1}{2}} (\sum_{\type=1}^\numSize \snumQuery_j (\tq_j - \hq_j)^2)^{\frac{1}{2}}\\
        & \leq \sqrt{\numPeriod} \cdot \frac{2\beta k_1}{\alpha} \sqrt{\numSize \log(\numPeriod)}\\
        & = \frac{2\beta k_1}{\alpha} \sqrt{\numSize \numPeriod \log(\numPeriod)}
    \end{aligned} 
\end{equation*}

Therefore, applying (\ref{eq:est_dj_1}), with probability at least $1-\frac{1}{\numPeriod}$,
\begin{equation}\label{eq:upper_bound_term2}
    \begin{aligned}
        \sum_{\type=1}^\numSize \mu_j |\tq_j - \hq_j| & \leq \sum_{\type=1}^\numSize \snumQuery_j |\tq_j - \hq_j| + \sum_{\type=1}^\numSize |\mu_j - \snumQuery_j| |\tq_j - \hq_j|\\
        & \leq \sum_{\type=1}^\numSize \snumQuery_j |\tq_j - \hq_j| + \sum_{\type=1}^\numSize |\mu_j - \snumQuery_j|\\
        & \leq \frac{2\beta k_1}{\alpha} \sqrt{\numSize \numPeriod \log(\numPeriod)} + 2 \sqrt{\numSize \numPeriod \log(\numPeriod)}
    \end{aligned}
\end{equation}

Combining (\ref{eq:upper_bound_term1}) and (\ref{eq:upper_bound_term2}), with probability at least $1-\frac{1}{\numPeriod}$, we have
\begin{equation*}
    \begin{aligned}
        \sum_{\type=1}^\numSize \mu_j |q_j^* - \hq_j| & \leq \sum_{\type=1}^\numSize \mu_j |q_j^* - \tq_j| + \sum_{\type=1}^\numSize \mu_j |\tq_j - \hq_j|\\
        & \leq \frac{4\beta}{\alpha} \log(\numPeriod) \sqrt{\numSize \numPeriod} + \frac{2\beta k_1 }{\alpha} \sqrt{\numSize \numPeriod \log(\numPeriod)} + 2 \sqrt{\numSize \numPeriod \log(\numPeriod)}\\
        & \leq (\frac{4\beta + 2\beta k_1 }{\alpha} + 2) \log(\numPeriod) \sqrt{\numSize \numPeriod}
    \end{aligned}
\end{equation*}

To conclude, considering (\ref{eq:offline_regret}) and (\ref{eq:Violation_offline}), we obtain the following upper bound for our regret:
\begin{equation*}
    \begin{aligned}
        \regret(\pi) & \leq \expect[\instance \sim G]{\sum_{j=1}^\numSize \mu_j \cdot (\uppreward_j + \uppreward \numResource \amax) \cdot |q_j^* - \hq_j|}\\
        & \leq (\uppreward + \uppreward \numResource \amax) \expect[\instance \sim G]{\sum_{j=1}^\numSize  \mu_j |q_j^* - \hq_j|}\\
        & \leq (\uppreward + \uppreward \numResource \amax)(\frac{4\beta + 2\beta k_1 }{\alpha} + 2) \log(\numPeriod) \sqrt{\numSize \numPeriod} + (\uppreward + \uppreward \numResource \amax)\uppreward \numPeriod \cdot \frac{1}{\numPeriod}\\
        & \leq (\uppreward + \uppreward \numResource \amax)(\frac{4\beta + 2\beta k_1 }{\alpha} + 2 + \uppreward) \log(\numPeriod) \sqrt{\numSize \numPeriod}\\
        & = O(\numResource \log(\numPeriod) \sqrt{\numSize \numPeriod})
    \end{aligned}
\end{equation*}
where $\uppreward, \amax, \alpha, \beta, k_1$ are all constants.

\section{Missing Proof in \texorpdfstring{\Cref{sec:online setting}}{Section 5}}
\label{apx:online}
\xhdr{Proof of \Cref{th:online}:}

We define $\numQuery_{j,t}$ and $\snumQuery_{j,t}$ as the indicator variables for the arrival of a query of $\type$ at time period $t$ in the primary problem instance and in historical sample stream respectively. Thus, both variables take values in $\{0,1\}$. It holds that $\expect[\instance \sim G]{\numQuery_{j,t}} = \expect[\instance \sim G]{\snumQuery_{j,t}} = \typedf_t(j)$. Furthermore, for each time period $t$ , exactly one type arrives, so we have $\sum_{j=1}^\numSize \numQuery_{j,t} = \sum_{j=1}^\numSize \snumQuery_{j,t} = 1$. Aggregating over time, we have $\sum_{t=1}^\numPeriod \numQuery_{j,t} = \numQuery_j, \sum_{t=1}^\numPeriod \snumQuery_{j,t} = \snumQuery_j$ for every $\type \in [\numSize]$.
With such definition of $\numQuery_{j,t}$ and $\snumQuery_{j,t}$, similar to \Cref{apx:offline}, we have
\begin{equation*}
    \begin{aligned}
        \regret(\pi) & \leq \expect[\instance \sim G]{\sum_{j=1}^\numSize \sum_{t=1}^\numPeriod \numQuery_{j,t} (\int_{1-q_j^*(\blambda^*)}^1 \cdfInverse_j(u)\, \dd u - \int_{1-q_{j,t}^{\pi}}^1 \cdfInverse_j(u)\, \dd u)}\\
        & \leq \expect[\instance \sim G]{\sum_{j=1}^\numSize \sum_{t=1}^\numPeriod \typedf_t(j) \cdot \uppreward_j \cdot |q_j^*(\blambda^*) - q_{j,t}^{\pi}|}\\
        & \leq \expect[\instance \sim G]{\sum_{j=1}^\numSize \sum_{t=1}^\numPeriod \typedf_t(j) \cdot \uppreward_j \cdot |q_j^*(\blambda^*) - q_{j,t}(\blambda_t)| +  \sum_{t=1}^\numPeriod V(\Hat{\bq}(\blambda_t))}
    \end{aligned}
\end{equation*}
where the second inequality holds because the reward for type $\type$ is upper bounded by $\uppreward_j$. $V(\Hat{\bq}(\blambda_t))$ denotes the penalty of constraint violation caused by directly applying $\{q_{j,t}(\blambda_t)\}_{j=1}^\numSize$ for time period $t$ in our algorithm.
Now we discuss the upper bound for $V(\Hat{\bq}(\blambda_t))$:
\begin{equation}\label{eq:Violation_online}
   \begin{aligned}
      V(\Hat{\bq}(\blambda_t)) & \leq \uppreward \sum_{i=1}^\numResource \max \{0, \sum_{j=1}^\numSize \typedf_t(j) a_{j,i} q_{j,t}(\blambda_t) - \sum_{j=1}^\numSize \typedf_t(j) a_{j,i} q_j^*(\blambda^*)\}\\
       & \leq \uppreward \sum_{i=1}^\numResource \sum_{j=1}^\numSize \typedf_t(j) a_{j,i} |q_j^*(\blambda^*) - q_{j,t}(\blambda_t)|\\
       & \leq \uppreward \numResource \amax \sum_{j=1}^\numSize \typedf_t(j) |q_j^*(\blambda^*) - q_{j,t}(\blambda_t)|
   \end{aligned} 
\end{equation}
where $\uppreward = \max_j \uppreward_j, \amax = \max_{j,i} a_{j,i}$.
Thus
\begin{equation*}
    \sum_{t=1}^\numPeriod V(\Hat{\bq}(\blambda_t)) \leq \uppreward \numResource \amax \sum_{j=1}^\numSize \sum_{t=1}^\numPeriod \typedf_t(j) |q_j^*(\blambda^*) - q_{j,t}(\blambda_t)|
\end{equation*}
\begin{equation}\label{eq:online_regret}
    \regret(\pi) \leq (\uppreward + \uppreward \numResource \amax) \cdot \expect[\instance \sim G]{\sum_{j=1}^\numSize \sum_{t=1}^\numPeriod \typedf_t(j) |q_j^*(\blambda^*) - q_{j,t}(\blambda_t)|)}
\end{equation}

For the next part of analysis, we discuss the upper bound for $\sum_{j=1}^\numSize \sum_{t=1}^\numPeriod \typedf_t(j) |q_j^*(\blambda^*) - q_{j,t}(\blambda_t)|$. 
We know that $\{q_j^*(\blambda^*)\}_{j=1}^\numSize$ in \ref{eq:q_optimal} is the optimal solution to (\ref{eq:V_fluid2}) and $\{q_{j,t}(\blambda_t)\}_{j=1}^\numSize$ in (\ref{eq:q_online}) is the optimal solution to \ref{eq:V_est_online}. For notational simplicity, we omit $\blambda$ and write $q_j^*(\blambda^*)$ as $q_j^*$, $q_{j,t}(\blambda_t)$ as $q_{j,t}$ in the following.

For each $j \in [\numSize]$, we define
\begin{equation*}
    G_j(q) = \int_{1-q}^1 \cdfInverse_j(u)\, \dd u
\end{equation*}

Following (\ref{eq:Gj_prop}), we know $G_j(q)$ is Lipschitz-continuous and $1/\beta$-strongly concave.
Similar to \Cref{apx:offline}, we introduce the intermediate estimation problem \ref{eq:V_inter_offline} and denote $\{\tq_j\}_{j=1}^\numSize$ as its optimal solution.
Then following the triangle inequality, we can decompose $|q_j^* - q_{j,t}|$ into two parts:
\begin{equation}\label{eq:inter_online}
    \begin{aligned}
        \sum_{j=1}^\numSize \sum_{t=1}^\numPeriod \typedf_t(j) |q_j^* - q_{j,t}| & \leq \sum_{j=1}^\numSize \sum_{t=1}^\numPeriod \typedf_t(j) |q_j^* - \tq_j| + \sum_{j=1}^\numSize \sum_{t=1}^\numPeriod \typedf_t(j) |\tq_j - q_{j,t}|\\
        & = \sum_{\type=1}^\numSize \mu_j |q_j^* - \tq_j| + \sum_{j=1}^\numSize \sum_{t=1}^\numPeriod \typedf_t(j) |\tq_j - q_{j,t}|
    \end{aligned}
\end{equation}
where the equality holds for $\mu_j = \expect[\instance \sim G]{\numQuery_j} = \sum_{t=1}^\numPeriod \typedf_t(j)$.

The upper bound for the first term $\sum_{\type=1}^\numSize \mu_j |q_j^* - \tq_j|$ follows directly from (\ref{eq:upper_bound_term1}). Now it remains to bound $\sum_{j=1}^\numSize \sum_{t=1}^\numPeriod \typedf_t(j) |\tq_j - q_{j,t}|$. We utilize the strongly concavity of $G_j$ for any $t \in [\numPeriod]$:
\begin{equation*}
    G_j (\tq_j) \leq G_j(q_{j,t}) + G_j'(q_{j,t})(\tq_j - q_{j,t}) - \frac{1}{2\beta}(\tq_j - q_{j,t})^2
\end{equation*}

For each time period $t$, we define
\begin{equation} \label{eq:Gjt_def}
    \hG_{j,t} (q) = \int_{1-q}^1 \estcdfInverse_{j,t}(u)\, \dd u
\end{equation}

Similar to (\ref{eq:optimal_cdf_gap_bound}), from the optimality of $\{\tq_j\}_{j=1}^\numSize$ to \ref{eq:V_inter_offline} and $\{q_{j,t}\}_{j=1}^\numSize$ to \ref{eq:V_est_online}, we have
\begin{equation*}
    \begin{aligned}
        \frac{1}{2\beta} \sum_{j=1}^\numSize \snumQuery_j (\tq_j - q_{j,t})^2 & \leq \sum_{j=1}^\numSize \snumQuery_j G_j(q_{j,t}) - \sum_{j=1}^\numSize \snumQuery_j G_j(\tq_j) + \sum_{j=1}^\numSize \snumQuery_j G_j'(q_{j,t})(\tq_j - q_{j,t})\\
        & \leq \sum_{j=1}^\numSize \snumQuery_j |\cdfInverse_j(1-q_{j,t}) - \estcdfInverse_{j,t}(1-q_{j,t})||\tq_j - q_{j,t}|\\
        & \leq \sum_{j=1}^\numSize \snumQuery_j \frac{\epsilon_{j,t}}{\alpha} |\tq_j - q_{j,t}|
    \end{aligned}
\end{equation*}
where  $\epsilon_{j,t} = \sup_x |\estcdf_{j,t}(x) - \cdf_j(x)|$.
By Cauchy-Schwarz Inequality,  we have
\begin{equation*}
    \begin{aligned}
        \frac{1}{2\beta} \sum_{j=1}^\numSize \snumQuery_j (\tq_j - q_{j,t})^2 & \leq \sum_{j=1}^\numSize \snumQuery_j \frac{\epsilon_{j,t}}{\alpha} |\tq_j - q_{j,t}|\\
        & \leq \frac{1}{\alpha} (\sum_{j=1}^\numSize \snumQuery_j (\tq_j - q_{j,t})^2)^{\frac{1}{2}} (\sum_{j=1}^\numSize \snumQuery_j (\epsilon_{j,t}^2))^{\frac{1}{2}}
    \end{aligned} 
\end{equation*}

We denote $n_{j,t}$ as the number of query type $\type$ arrived from time period 1 to $t$. Under \Cref{assump:MAP}, it follows that $\expect[\instance \sim G]{n_{j,t}} \geq \gamma \cdot t$ and $\expect[\instance \sim G]{\snumQuery_j} \geq \gamma \cdot \numPeriod$. By Chernoff bound, we know that with probabilty at least $1-\delta$, 
\begin{equation*}
    \snumQuery_j \geq \expect[\instance \sim G]{\snumQuery_j} - \sqrt{2\expect[\instance \sim G]{\snumQuery_j} \log(\frac{1}{\delta})}, \quad n_{j,t} \geq \expect[\instance \sim G]{n_{j,t}} - \sqrt{2\expect[\instance \sim G]{n_{j,t}} \log(\frac{1}{\delta})}
\end{equation*}

Therefore when $t \geq t_0 = \frac{8}{\gamma}\log(\numPeriod)$, with probability at least $1 - \frac{1}{\numPeriod}$, we have $\snumQuery_j \geq \frac{\gamma \numPeriod}{2}$ and $n_{j,t} \geq \frac{\gamma t}{2}$.
By \Cref{le:est_Fj}, with probability at least $1 - \frac{1}{\numPeriod}$, we have $\epsilon_{j,t} = \sup_x |\estcdf_{j,t}(x) - \cdf_j(x)| \leq k_1 \sqrt{\log (\numPeriod) / n_{j,t}}$, where $k_1$ is a constant.
Thus with probability at least $1-\frac{2}{\numPeriod}$,
\begin{equation*}
    \frac{\gamma \numPeriod}{2} \sum_{j=1}^\numSize (\tq_j - q_{j,t})^2 \leq \sum_{j=1}^\numSize \snumQuery_j (\tq_j - q_{j,t})^2 \leq (\frac{2\beta}{\alpha})^2 \sum_{j=1}^\numSize \snumQuery_j (\epsilon_{j,t}^2) \leq (\frac{2\beta}{\alpha})^2 \sum_{j=1}^\numSize \snumQuery_j \cdot \frac{2 k_1^2 \log(\numPeriod)}{\gamma t}
\end{equation*}

Consequently,
\begin{equation*}
    \sum_{j=1}^\numSize (\tq_j - q_{j,t})^2 \leq \frac{2}{\gamma \numPeriod} (\frac{2\beta}{\alpha})^2 \sum_{j=1}^\numSize \snumQuery_j \cdot \frac{2 k_1^2 \log(\numPeriod)}{\gamma t} \leq (\frac{4\beta k_1}{\alpha \gamma})^2 \frac{\log(\numPeriod)}{t}
\end{equation*}

Therefore, by Cauchy-Schwarz Inequality, with probability at least $1-\frac{2}{\numPeriod}$,
\begin{equation}\label{eq:upper_bound_term2_online}
    \begin{aligned}
        \sum_{j=1}^\numSize \sum_{t=1}^\numPeriod \typedf_t(j) |\tq_j - q_{j,t}| & \leq \sum_{t=1}^\numPeriod \sum_{j=1}^\numSize |\tq_j - q_{j,t}| \\
        & \leq \sqrt{\numSize} (\frac{4\beta k_1}{\alpha \gamma} \sum_{t=1}^\numPeriod \sqrt{\frac{\log(\numPeriod)}{t}} + 2 t_0)\\
        & \leq \frac{8\beta k_1}{\alpha \gamma} \sqrt{\log(\numPeriod) \cdot \numSize \numPeriod} + \frac{16}{\gamma} \sqrt{\numSize} \log(\numPeriod)
    \end{aligned}
\end{equation}

Combining (\ref{eq:inter_online}), (\ref{eq:upper_bound_term1}) and (\ref{eq:upper_bound_term2_online}), with probability at least $1-\frac{2}{\numPeriod}$, we have
\begin{equation} \label{eq:inter_online2}
    \begin{aligned}
        \sum_{j=1}^\numSize \sum_{t=1}^\numPeriod \typedf_t(j) |q_j^* - q_{j,t}| & \leq \sum_{\type=1}^\numSize \mu_j |q_j^* - \tq_j| + \sum_{j=1}^\numSize \sum_{t=1}^\numPeriod \typedf_t(j) |\tq_j - q_{j,t}|\\ 
        & \leq \frac{4\beta}{\alpha} \log(\numPeriod) \sqrt{\numSize \numPeriod} + \frac{8\beta k_1}{\alpha \gamma} \sqrt{\log(\numPeriod) \cdot \numSize \numPeriod} + \frac{16}{\gamma} \sqrt{\numSize} \log(\numPeriod)\\
        & \leq (\frac{4\beta}{\alpha} + \frac{8\beta k_1}{\alpha \gamma} + \frac{16}{\gamma}) \log(\numPeriod) \sqrt{\numSize \numPeriod}
    \end{aligned}
\end{equation}

Finally, considering (\ref{eq:inter_online2}) and (\ref{eq:online_regret}) we obtain the following upper bound for online regret:
\begin{equation*}
    \begin{aligned}
        \regret(\pi) & \leq (\uppreward + \uppreward \numResource \amax) \cdot \expect[\instance \sim G]{\sum_{j=1}^\numSize \sum_{t=1}^\numPeriod \typedf_t(j) |q_j^* - q_{j,t}|)}\\
        & \leq (\uppreward + \uppreward \numResource \amax)\left(\frac{4\beta}{\alpha} + \frac{8\beta k_1}{\alpha \gamma} + \frac{16}{\gamma}\right) \log(\numPeriod) \sqrt{\numSize \numPeriod} + (\uppreward + \uppreward \numResource \amax)\uppreward \numPeriod \cdot \frac{2}{\numPeriod}\\
        & \leq (\uppreward + \uppreward \numResource \amax)\left(\frac{4\beta}{\alpha} + \frac{8\beta k_1}{\alpha \gamma} + \frac{16}{\gamma} + 2\uppreward\right) \log(\numPeriod) \sqrt{\numSize \numPeriod}\\
        & = O(\numResource \log(\numPeriod) \sqrt{\numSize \numPeriod})
    \end{aligned}
\end{equation*}
where $\uppreward, \amax, \alpha, \beta, \gamma, k_1$ are all constants.

\section{Further Discussion and Missing Proof in \texorpdfstring{\Cref{sec:poly-log regret}}{Section 6}} 
\label{apx:log}
\xhdr{Introduction to Semi-fluid Relaxation:}

In order to obtain a poly-logarithmic regret with our policy in \Cref{sec:poly-log regret}, we now introduce the definition of \emph{semi-fluid} relaxation. For a fixed instance $\instance$ with $\bnumQuery = (\numQuery_{1}, \cdots, \numQuery_{\numSize})$, we formulate the semi-fluid relaxation of \ref{eq:V_off}:
\begin{align}
    \label{eq:V_semi} 
    \tag{$\Vsemi_{\bconstraint}(\instance)$}
    \arraycolsep=5.4pt\def\arraystretch{1.7}
    \begin{array}{llll}
    \max\limits_{\xbf} ~ &
    \displaystyle
    \sum\nolimits_{j\in[\numSize]} \numQuery_j \cdot \expect[\reward \sim \cdf_j]{\reward \cdot x_j(\reward)}
    &
    \text{s.t.}
    \\
    &
    \displaystyle\sum\nolimits_{j\in[\numSize]} \numQuery_j \cdot a_{j,i} \cdot \expect[\reward \sim \cdf_j]{x_j(\reward)} \leq \constraint_i
    & i\in[\numResource]
    \\
    & x_j(\reward) \in [0,1] & j \in [\numSize], \reward \in [\lowreward_j, \uppreward_j]
    \end{array}
\end{align}
where $\numQuery_j$ denote the number of type $\type$ query arrivals and $\bnumQuery$ depends on the sample path $\instance$. From \Cref{le:Vfluid_Voff}, we know that \emph{semi-fluid} relaxation also implies an upper bound for offline optimum as $\Vfluid_{\bconstraint} = \expect[\instance \sim G]{\Vsemi_{\bconstraint}(\instance)} \geq \expect[\instance \sim G]{\Voff_{\bconstraint}(\instance)}$.

Following \Cref{le:threshold_property}, the \emph{semi-fluid} relaxation problem can be equivalently rewritten as:
\begin{align}
    \label{eq:V_semi2} 
    \tag{$\bVsemi_{\bconstraint}(\instance)$}
    \arraycolsep=5.4pt\def\arraystretch{1.7}
    \begin{array}{llll}
    \max\limits_{\bq} ~ &
    \displaystyle
    \sum\nolimits_{j\in[\numSize]} \numQuery_j \cdot \int_{1-q_j}^1 \cdfInverse_j(u)\, \dd u
    &
    \text{s.t.}
    \\
    &
    \displaystyle\sum\nolimits_{j\in[\numSize]} \numQuery_j \cdot a_{j,i} \cdot q_j \leq \constraint_i
    & i\in[\numResource]
    \\
    & q_j \in [0,1] & j \in [\numSize]
    \end{array}
\end{align}
where the decision variable $q_j$ represents the probability of serving a query of type $\type$. 

To achieve poly-logarithmic regret, we now consider the problem starting from time period $t$. We define random variable for type $\type$ query arrivals from period $t$ to $\numPeriod$ as $\tnumQuery_{j,t}$ in the problem instance $\instance$. Then on the remaining problem instance $\instance_t = \{(\reward_s, \size_s)\}_{s=t}^\numPeriod$, we denote the following semi-fluid problem at time period $t$, which serves as a relaxation of the total reward collected by the prophet from time period $t$ to $\numPeriod$ given the remaining capacity $\brconstraint$.
\begin{align}
    \label{eq:V_semi_t}
    \tag{$\bVsemi_{\brconstraint}(\instance_t)$}
    \arraycolsep=5.4pt\def\arraystretch{1.7}
    \begin{array}{llll}
    \max\limits_{\bq} ~ &
    \displaystyle
    \sum\nolimits_{j\in[\numSize]} \tnumQuery_{j,t} \cdot \int_{1-q_j}^1 \cdfInverse_j(u)\, \dd u 
    &
    \text{s.t.}
    \\
    &
    \displaystyle\sum\nolimits_{j\in[\numSize]} \tnumQuery_{j,t} \cdot a_{j,i} \cdot q_j \leq \rconstraint_i
    & i\in[\numResource]
    \\
    & q_j \in [0,1] & j \in [\numSize]
    \end{array}
\end{align}
and we denote $\{\tq_{\type,t}^*\}_{j=1}^{\numSize}$ as its optimal solution.

\xhdr{Proof of \Cref{le:q_gap}:}

We introduce an intermediate variable $\tq_{\type,t}$ and decompose $|\tq_{\type,t}^*-\hq_{\type,t}|$ into two parts using the triangle inequality. Specifically, for instance $\instance_t$, we consider an intermediate estimation problem in which the reward distributions are known accurately:
\begin{align}
    \label{eq:V_inter_log}
    \tag{$\tV_{t,\brconstraint}(\instance_t)$}
    \arraycolsep=5.4pt\def\arraystretch{1.7}
    \begin{array}{llll}
    \max\limits_{\bq} ~ &
    \displaystyle
    \sum\nolimits_{j\in[\numSize]} \tsnumQuery_{j,t} \cdot \int_{1-q_j}^1 \cdfInverse_j(u)\, \dd u 
    &
    \text{s.t.}
    \\
    &
    \displaystyle\sum\nolimits_{j\in[\numSize]} \tsnumQuery_{j,t} \cdot a_{j,i} \cdot q_j \leq \rconstraint_i
    & i\in[\numResource]
    \\
    & q_j \in [0,1] & j \in [\numSize]
    \end{array}
\end{align}
and let $\{\tq_{\type,t}\}_{j=1}^\numSize$ denote its optimal solution. Then we have that for any $\type \in [\numSize]$,
\begin{equation} \label{eq:q_gap_log}
    |\tq_{\type,t}^*-\hq_{\type,t}| \leq |\tq_{\type,t}^*-\tq_{\type,t}| + |\tq_{\type,t}-\hq_{\type,t}|
\end{equation}

We first consider the first term on the right-hand side.
Following the strong concavity in (\ref{eq:Gj_prop}), we have
\begin{equation*}
    \frac{1}{\beta}(\tq_{\type,t} - \tq_{\type,t}^*)^2 \leq  \left(G_j'(\tq_{\type,t})-G_j'(\tq_{\type,t}^*)\right)(\tq_{\type,t}^* - \tq_{\type,t})
\end{equation*}

Following the analysis in \Cref{apx:offline} by substituting $\mu_j$ and $\constraint_i$ for $\tnumQuery_{j,t}$ and $\rconstraint_i$ for every $j$ and $i$, we have
\begin{equation*}
    \frac{1}{\beta} \sum_{\type=1}^\numSize \tnumQuery_{j,t} (\tq_{\type,t} - \tq_{\type,t}^*)^2 \leq \frac{1}{\alpha} \sum_{\type=1}^\numSize |\tnumQuery_{j,t} - \tsnumQuery_{j,t}||\tq_{\type,t} - \tq_{\type,t}^*|
\end{equation*}

By Cauchy-Schwarz Inequality,  we have
\begin{equation*}
    \begin{aligned}
        \frac{1}{\beta} \sum_{\type=1}^\numSize \tnumQuery_{j,t} (\tq_{\type,t} - \tq_{\type,t}^*)^2 & \leq \frac{1}{\alpha} \sum_{\type=1}^\numSize |\tnumQuery_{j,t} - \tsnumQuery_{j,t}||\tq_{\type,t} - \tq_{\type,t}^*|\\
        & \leq \frac{1}{\alpha} (\sum_{\type=1}^\numSize \tnumQuery_{j,t} (\tq_{\type,t} - \tq_{\type,t}^*)^2)^{\frac{1}{2}} (\sum_{\type=1}^\numSize \frac{|\tnumQuery_{j,t} - \tsnumQuery_{j,t}|^2}{\tnumQuery_{j,t}})^{\frac{1}{2}}
    \end{aligned} 
\end{equation*}

Therefore,
\begin{equation*} 
    \sum_{\type=1}^\numSize \tnumQuery_{j,t} (\tq_{\type,t} - \tq_{\type,t}^*)^2 \leq (\frac{\beta}{\alpha})^2 \sum_{\type=1}^\numSize \frac{|\tnumQuery_{j,t} - \tsnumQuery_{j,t}|^2}{\tnumQuery_{j,t}}
\end{equation*}

From \Cref{apx:est_arrival-distribution}, we have (\ref{eq:est_dj_2}). Since $\tnumQuery_{j,t}$ and $\tsnumQuery_{j,t}$ are independent identically distributed, with probability at least $1-\frac{1}{\numPeriod}$, we have
\begin{equation*}
    \begin{aligned}
        \sum_{\type=1}^\numSize \tnumQuery_{j,t} |\tq_{\type,t} - \tq_{\type,t}^*| & \leq (\sum_{\type=1}^\numSize \tnumQuery_{j,t} )^{\frac{1}{2}} (\sum_{\type=1}^\numSize \tnumQuery_{j,t} (\tq_{\type,t} - \tq_{\type,t}^*)^2)^{\frac{1}{2}}\\
        & \leq \sqrt{s} \cdot \frac{\beta}{\alpha} (\sum_{\type=1}^\numSize \frac{|\tnumQuery_{j,t} - \tsnumQuery_{j,t}|^2}{\tnumQuery_{j,t}})^{\frac{1}{2}}\\
        & \leq \sqrt{s} \cdot \frac{\beta}{\alpha} \cdot 4\sqrt{2} \log(2\numPeriod) \sqrt{\numSize}\\
        & \leq \frac{4\beta}{\alpha} \log\numPeriod \sqrt{\numSize s}
    \end{aligned} 
\end{equation*}
where the first inequality holds for Cauchy-Schwarz Inequality and the second holds for $\sum_{\type=1}^\numSize \tnumQuery_{j,t} = s$.

Now that we have $\tnumQuery_{j,t} |\tq_{\type,t} - \tq_{\type,t}^*| \leq \frac{4\beta}{\alpha} \log\numPeriod \sqrt{\numSize s}$ holds for any $\type \in [\numSize]$. With \Cref{assump:MAP}, we know that $\tnumQuery_{j,t} \geq \gamma \cdot s$, thus with probability at least $1-\frac{1}{\numPeriod}$,
\begin{equation} \label{eq:q_gap_log_term1}
    |\tq_{\type,t} - \tq_{\type,t}^*| \leq \frac{4 \beta \sqrt{\numSize}}{\alpha \gamma} \frac{\log \numPeriod}{\sqrt{s}}, \forall \type \in [\numSize]
\end{equation}

For the second term on the right-hand side of (\ref{eq:q_gap_log}), we follow the definition of $\hG_{\type,t}$ in (\ref{eq:Gjt_def}) and apply the strong concavity of $G_j$.
Consequently, we have
\begin{equation*}
    \begin{aligned}
        \frac{1}{2\beta} \sum_{j=1}^\numSize \tsnumQuery_{j,t} (\tq_{\type,t} - \hq_{\type,t})^2 & \leq \sum_{j=1}^\numSize \tsnumQuery_{j,t} G_j(\hq_{\type,t}) - \sum_{j=1}^\numSize \tsnumQuery_{j,t} G_j(\tq_{\type,t}) + \sum_{j=1}^\numSize \tsnumQuery_{j,t} G_j'(\hq_{\type,t})(\tq_{\type,t} - \hq_{\type,t})\\
        & \leq \sum_{j=1}^\numSize \tsnumQuery_{j,t} G_j'(\hq_{\type,t})(\tq_{\type,t} - \hq_{\type,t})\\
        & = \sum_{j=1}^\numSize \tsnumQuery_{j,t} \hG_{j,t}'(\hq_{\type,t})(\tq_{\type,t} - \hq_{\type,t}) + \sum_{j=1}^\numSize \tsnumQuery_{j,t} (G_j'(\hq_{\type,t})-\hG_{j,t}'(\hq_{\type,t}))(\tq_{\type,t} - \hq_{\type,t})\\
        & \leq \sum_{j=1}^\numSize \tsnumQuery_{j,t} |\cdfInverse_j(1-\hq_{j,t}) - \estcdfInverse_{j,t}(1-\hq_{j,t})||\tq_{\type,t} - \hq_{\type,t}|\\
        & \leq \sum_{j=1}^\numSize \tsnumQuery_{j,t} \frac{\epsilon_{j,t}}{\alpha} |\tq_{\type,t} - \hq_{\type,t}|
    \end{aligned}
\end{equation*}
where  $\epsilon_{j,t} = \sup_x |\estcdf_{j,t}(x) - \cdf_j(x)|$. The second inequality relies on the optimality of $\{\tq_{\type,t}\}_{j=1}^\numSize$ to \ref{eq:V_inter_log}, and the third inequality holds for the optimality of $\{\hq_{j,t}\}_{j=1}^\numSize$ to \ref{eq:V_est_log_t}.

By Cauchy-Schwarz Inequality,  we have
\begin{equation*}
    \begin{aligned}
        \frac{1}{2\beta} \sum_{j=1}^\numSize \tsnumQuery_{j,t} (\tq_{\type,t} - \hq_{\type,t})^2 & \leq \sum_{j=1}^\numSize \tsnumQuery_{j,t} \frac{\epsilon_{j,t}}{\alpha} |\tq_{\type,t} - \hq_{\type,t}|\\
        & \leq \frac{1}{\alpha} (\sum_{j=1}^\numSize \tsnumQuery_{j,t} (\tq_{\type,t} - \hq_{\type,t})^2)^{\frac{1}{2}} (\sum_{j=1}^\numSize \tsnumQuery_{j,t} (\epsilon_{j,t}^2))^{\frac{1}{2}}
    \end{aligned} 
\end{equation*}

By \Cref{le:est_Fj}, with probability at least $1 - \delta$, we have $\epsilon_{j,t} = \sup_x |\estcdf_{j,t}(x) - \cdf_j(x)| \leq k_1 \sqrt{\log (1/\delta) / n_{j,t}}$, where $k_1$ is a constant.
Moreover, under \Cref{assump:MAP2}, we have $\expect[\instance \sim G]{n_{j,t}} \geq \gamma t = \gamma (\numPeriod-s+1)$. 
Therefore, with probability at least $1 - \frac{1}{\numPeriod}$, we have 
\begin{equation*}
    \begin{aligned}
        \sum_{j=1}^\numSize \tsnumQuery_{j,t} (\tq_{\type,t}-\hq_{\type,t})^2 & \leq (\frac{2\beta}{\alpha})^2 \cdot \sum_{j=1}^\numSize \tsnumQuery_{j,t} (\epsilon_{j,t}^2)\\
        & \leq (\frac{2\beta}{\alpha})^2 \cdot \frac{k_1^2 \log \numPeriod}{\gamma (\numPeriod-s+1)} \sum_{j=1}^\numSize \tsnumQuery_{j,t}\\
        & = (\frac{2\beta}{\alpha})^2 \cdot \frac{k_1^2 s \log \numPeriod}{\gamma(\numPeriod-s+1)}
    \end{aligned}
\end{equation*}
where the last equality holds for $\sum_{\type=1}^\numSize \tsnumQuery_{j,t} = s$.
Therefore by Cauchy-Schwarz Inequality,
\begin{equation*}
    \begin{aligned}
        \sum_{j=1}^\numSize \tsnumQuery_{j,t} |\tq_{\type,t}-\hq_{\type,t}| 
        & \leq (\sum_{\type=1}^\numSize \tsnumQuery_{j,t} )^{\frac{1}{2}} (\sum_{\type=1}^\numSize \tsnumQuery_{j,t} (\tq_{\type,t}-\hq_{\type,t})^2)^{\frac{1}{2}}\\
        & \leq s \cdot \frac{2\beta k_1 }{\alpha} \sqrt{\frac{\log\numPeriod}{\gamma(\numPeriod-s+1)}}
    \end{aligned} 
\end{equation*}

Now that we have $\tsnumQuery_{j,t} |\tq_{\type,t}-\hq_{\type,t}| \leq s \cdot \frac{2\beta k_1 }{\alpha} \sqrt{\frac{\log\numPeriod}{\gamma(\numPeriod-s+1)}}$ holds for any $\type \in [\numSize]$. With \Cref{assump:MAP}, we know that $\tsnumQuery_{j,t} \geq \gamma \cdot s$, thus with probability at least $1-\frac{1}{\numPeriod}$,
\begin{equation} \label{eq:q_gap_log_term2}
    |\tq_{\type,t}-\hq_{\type,t}| \leq \frac{2\beta k_1 }{\alpha \sqrt{\gamma^3}} \sqrt{\frac{\log\numPeriod}{\numPeriod-s+1}}, \forall \type \in [\numSize]
\end{equation}

Combining (\ref{eq:q_gap_log}), (\ref{eq:q_gap_log_term1}) and (\ref{eq:q_gap_log_term2}), for any $\type \in [\numSize]$, with probability at least $1-\frac{1}{\numPeriod}$,
\begin{equation*}
    \begin{aligned}
        |\tq_{\type,t}^*-\hq_{\type,t}| & \leq \frac{4\beta \sqrt{\numSize}}{\alpha \gamma} \frac{\log \numPeriod}{\sqrt{s}} + \frac{2\beta k_1 }{\alpha \sqrt{\gamma^3}} \sqrt{\frac{\log\numPeriod}{\numPeriod-s+1}}\\
        & \leq (\frac{4\beta \sqrt{\numSize}}{\alpha \gamma} + \frac{2\beta k_1 }{\alpha \sqrt{\gamma^3}}) (\frac{\log \numPeriod}{\sqrt{s}} + \frac{\log \numPeriod}{\sqrt{\numPeriod-s+1}})
    \end{aligned}
\end{equation*}
where $\alpha, \beta, \gamma, k_1, \uppreward$ are all constants.

\xhdr{Proof of \Cref{th:log}:}

We denote by $\bV_{\brconstraint}(\instance_t)$ a relaxation of the total reward collected by the prophet from time period $t$ to $\numPeriod$ given the remaining problem instance $\instance_t = \{(\reward_s, \size_s)\}_{s=t}^\numPeriod$ and the remaining capacity $\brconstraint$.
The regret of any online policy $\pi$ over the whole time horizon $\numPeriod$ can therefore be upper bound by the expected gap between $\bV_{\bconstraint}(\instance_1)$ and $V_{\bconstraint}^{\pi}(\instance)$, where $\bconstraint=(\constraint_{1}, \cdots, \constraint_{\numResource})$ is a vector of initial capacity for all resources, i.e.,
\begin{equation} \label{eq:regret_relaxation}
    \regret(\pi) \leq \expect[\instance_1 \sim G]{\bV_{\bconstraint}(\instance_1)} - \expect[\instance \sim G]{V^{\pi}(\instance)}
\end{equation}

For each $t \in [\numPeriod]$, we denote $\brconstraint_t^{\pi}=(\rconstraint_{t,1}^{\pi}, \cdots, \rconstraint_{t,\numResource}^{\pi})$ as the remaining capacities of the resources at the beginning of a period $t$ during the implementation of policy $\pi$ and $\brconstraint_t^{\pi}$ is random for each period $t$ because of the randomness in the problem instance $\instance$ and policy $\pi$. Note that $\brconstraint_{1}^{\pi} = \bconstraint$ and $\bV_{\brconstraint}(\instance_{\numPeriod+1})=0$ for every $\brconstraint$, the regret upper bound in (\ref{eq:regret_relaxation}) can be decomposed as:
\begin{equation*}
    \begin{aligned}
        \expect[\instance_1 \sim G]{\bV_{\bconstraint}(\instance_1)} - \expect[\instance \sim G]{V^{\pi}(\instance)} & = \expect[\instance_1 \sim G]{\sum_{t=1}^{\numPeriod} (\bV_{\brconstraint_t^{\pi}}(\instance_t) - \bV_{\brconstraint_{t+1}^{\pi}}(\instance_{t+1}))} - \expect[\instance \sim G]{V^{\pi}(\instance)}\\
        & = \sum_{t=1}^{\numPeriod} \expect[\instance_t \sim G]{\bV_{\brconstraint_t^{\pi}}(\instance_t) - \bV_{\brconstraint_{t+1}^{\pi}}(\instance_{t+1}) - \reward_t \cdot \piDecision_t}\\
        & = \sum_{t=1}^{\numPeriod} \expect[\instance_t \sim G]{\bV_{\brconstraint_t^{\pi}}(\instance_t) - \bV_{\brconstraint_{t}^{\pi} - \size_t \cdot \piDecision_t}(\instance_{t+1}) - \reward_t \cdot \piDecision_t}
    \end{aligned}
\end{equation*}
where $\piDecision_t$ is the decision for period $t$ in the policy $\pi$.
For each $\brconstraint \geq 0$, we define the \emph{myopic} regret:
\begin{equation} \label{eq:myopic_regret}
    \myopic_t(\pi,\brconstraint) = \expect[\instance_t \sim G]{\bV_{\brconstraint}(\instance_t) - \bV_{\brconstraint - \size_t \cdot \piDecision_t}(\instance_{t+1}) - \reward_t \cdot \piDecision_t}
\end{equation}

In practice, applying \Cref{alg:log} may lead to infeasibility, which introduces additional regret between our policy and the benchmark. To illustrate this, consider a virtual buffer that is introduced for the algorithm, containing exactly one unit of resource to be consumed by an arriving query. When the real remaining capacity is sufficient for accepting the query, the policy behaves identically in both the real and the virtual settings. However, once the real remaining capacity is no longer enough to accept an arriving query, the policy will reject all subsequent queries in the real setting. Consequently, the only discrepancy between the virtual and real cases occurs at the first time period when the real resource capacity runs out: in the virtual case, the query can still be accepted using the remaining buffer, whereas in practice it must be rejected. This discrepancy results in a performance gap of at most $\uppreward$ for each resource $i \in [\numResource]$. Therefore, the total regret can be rewritten as follows:
\begin{equation} \label{eq:myopic_decompose}
    \regret(\pi) \leq  \sum_{t=1}^{\numPeriod} \expect[\brconstraint_t^{\pi}]{\myopic_t(\pi,\brconstraint_t^{\pi})} + \numResource \cdot \uppreward
\end{equation}

We now proceed to offer an upper bound for myopic regret. As defined in \ref{eq:V_semi_t}, we have
\begin{equation}
    \bVsemi_{\brconstraint}(\instance_t) = \sum_{j=1}^\numSize \tnumQuery_{j,t+1} \cdot \int_{1-\tq_{\type,t}^*}^1 \cdfInverse_{\type}(u)\, \dd u + \int_{1-\tq_{\type_t,t}^*}^1 \cdfInverse_{\type_t}(u)\, \dd u
\end{equation}
where we denote by $\type_t$ the type of query $t$ in the instance $\instance$. Set $\bV_{\brconstraint}(\instance_t):=\bVsemi_{\brconstraint}(\instance_t)$ in (\ref{eq:myopic_regret}) and denote $q_{\type,t}^{\pi}$ as the probability that query $\query$ will be served by the online policy $\pi$:
\begin{equation} \label{eq:myopic_regret_1}
    \begin{aligned}
        \myopic_t(\pi,\brconstraint) & = \sexpect[\instance_{t+1} \sim G]{\sum_{j=1}^\numSize \tnumQuery_{j,t+1} \cdot \int_{1-\tq_{\type,t}^*}^1 \cdfInverse_{\type}(u)\, \dd u + \int_{1-\tq_{\type_t,t}^*}^1 \cdfInverse_{\type_t}(u)\, \dd u \\
        & \quad - \int_{1-q_{\type_t,t}^{\pi}}^1 \cdfInverse_{\type_t}(u)\, \dd u - \sexpect[\reward \sim \cdf_{\type_t}]{\bVsemi_{\brconstraint - \size_{\type_t} \cdot \piDecision_t(\reward)}(\instance_{t+1})}} \\
        & = \sexpect[\instance_{t+1} \sim G]{\sum_{j=1}^\numSize \tnumQuery_{j,t+1} \cdot \int_{1-\tq_{\type,t}^*}^1 \cdfInverse_{\type}(u)\, \dd u + \int_{1-\tq_{\type_t,t}^*}^{1-q_{\type_t,t}^{\pi}} \cdfInverse_{\type_t}(u)\, \dd u \\
        & \quad - q_{\type_t,t}^{\pi} \bVsemi_{\brconstraint - \size_{\type_t}}(\instance_{t+1}) -(1-q_{\type_t,t}^{\pi}) \bVsemi_{\brconstraint}(\instance_{t+1})}
    \end{aligned}
\end{equation}

We now discuss three cases dependent on the different computed quantile values from \ref{eq:V_est_log_t} and construct feasible solution to $\bVsemi_{\brconstraint - \size_{\type_t}}(\instance_{t+1})$ and $\bVsemi_{\brconstraint}(\instance_{t+1})$ to upper bound the myopic regret.

\textbf{Case 1:} when $\hq_{\type_t,t} \geq 1-2\kappa (\frac{\log \numPeriod}{\sqrt{s}}+\frac{\log \numPeriod}{\sqrt{\numPeriod-s+1}})$.
From \Cref{le:q_gap}, we have 
\begin{equation*}
    |\tq_{\type_t,t}^*-\hq_{\type_t,t}| \leq \kappa (\frac{\log \numPeriod}{\sqrt{s}} + \frac{\log \numPeriod}{\sqrt{\numPeriod-s+1}})
\end{equation*}
It implies that $\tq_{\type_t,t}^* \geq 1-3\kappa (\frac{\log \numPeriod}{\sqrt{s}}+\frac{\log \numPeriod}{\sqrt{\numPeriod-s+1}}) \geq \frac{1}{2}$ when $s \geq 144\kappa^2 (\log \numPeriod)^2$ and $\numPeriod-s+1 \geq 144\kappa^2 (\log \numPeriod)^2$.
We know that
\begin{equation*}
    \brconstraint \geq \tsnumQuery_{\type_t,t} \cdot \hq_{\type_t,t} \cdot \size_{\type_t} \geq \gamma \cdot s \cdot \frac{\size_{\type_t}}{2} \geq \size_{\type_t}
\end{equation*}
for large $s \geq \frac{2}{\gamma}$. Therefore, we always have enough remaining capacity to serve quert $t$ with type $\type_t$.
We know that query $t$ of type $\type_t$ should be accepted by our algorithm in a high quantile. For simplicity, we set $q_{\type_t,t}^{\pi} =1$. Thus we only need to construct a feasible solution to $\bVsemi_{\brconstraint - \size_{\type_t}}(\instance_{t+1})$ as it contributes negatively to our myopic regret.

From the feasibility of $\{\tq_{\type,t}^*\}_{\type=1}^{\numSize}$, we know
\begin{equation} \label{eq:qt_constraint}
    \sum_{\type=1}^{\numSize} \tnumQuery_{\type,t+1} \cdot a_{j,i} \cdot \tq_{\type,t}^* + a_{\type_t,i} \cdot\tq_{\type_t,t}^* \leq \rconstraint_i, \forall i \in [\numResource]
\end{equation}

We construct the following solution $\{\tq_{\type,t}'\}_{\type=1}^{\numSize}$ satisfying
\begin{equation} \label{eq:q'}
    \tq_{\type,t}' = \tq_{\type,t}^*, \forall \type \neq \type_t \quad \text{and} \quad \tq_{\type_t,t}' = \tq_{\type_t,t}^* + \frac{\tq_{\type_t,t}^*-1}{\tnumQuery_{\type_t,t+1}}
\end{equation}

Then we have $\sum_{\type=1}^{\numSize} \tnumQuery_{\type,t+1} \cdot a_{j,i} \cdot \tq_{\type,t}' \leq \rconstraint_i - a_{\type_t,i}$ for every $i \in [\numResource]$. Therefore $\{\tq_{\type,t}'\}_{\type=1}^{\numSize}$ is a feasible solution to $\bVsemi_{\brconstraint - \size_{\type_t}}(\instance_{t+1})$, where $\tq_{\type_t,t}' \geq 0$ follows from $\tq_{\type_t,t}^* \geq \frac{1}{2}$ and $\tnumQuery_{\type_t,t+1} \geq 1$.

Consequently, following (\ref{eq:myopic_regret_1}), setting $q_{\type_t,t}^{\pi}=1$, we have an upper bound for myopic regret:
\begin{equation*} 
    \begin{aligned}
        & \myopic_t(\pi,\brconstraint)\\
        \leq & \expect[\instance_{t+1} \sim G]{\sum_{j=1}^\numSize \tnumQuery_{j,t+1} \cdot \int_{1-\tq_{\type,t}^*}^1 \cdfInverse_{\type}(u)\, \dd u + \int_{1-\tq_{\type_t,t}^*}^{1-q_{\type_t,t}^{\pi}} \cdfInverse_{\type_t}(u)\, \dd u  - \sum_{j=1}^\numSize \tnumQuery_{j,t+1} \cdot \int_{1-\tq_{\type,t}'}^1 \cdfInverse_{\type}(u)\, \dd u }\\
        = & \expect[\instance_{t+1} \sim G]{\tnumQuery_{\type_t,t+1} \cdot \int_{1-\tq_{\type_t,t}^*}^{1-\tq_{\type_t,t}'} \cdfInverse_{\type_t}(u)\, \dd u + \int_{1-\tq_{\type_t,t}^*}^{1-q_{\type_t,t}^{\pi}} \cdfInverse_{\type_t}(u)\, \dd u}
    \end{aligned}
\end{equation*}

We introduce the following lemma:
\begin{lemma} \label{le:bound_intF}
    For any $q_1, q_2 \in [0,1]$, it holds that
    \begin{equation*}
        \int_{1-q_1}^{1-q_2} \cdfInverse_j(u)\, \dd u \leq \cdfInverse_j(1-q_1)\cdot (q_1-q_2) + \frac{(q_1-q_2)^2}{2\alpha}
    \end{equation*}
    for any $\type \in [\numSize]$, where $\alpha$ is the lower bound of density function $\pdf_j(\cdot)$ defined in \Cref{assump:reward}.
\end{lemma}

Applying \Cref{le:bound_intF}, we have
\begin{equation*}
    \begin{aligned}
        & \int_{1-\tq_{\type_t,t}^*}^{1-\tq_{\type_t,t}'} \cdfInverse_{\type_t}(u)\, \dd u \leq \cdfInverse_{\type_t}(1-\tq_{\type_t,t}^*) \cdot \frac{1-\tq_{\type_t,t}^*}{\tnumQuery_{\type_t,t+1}} + \frac{(1-\tq_{\type_t,t}^*)^2}{2\alpha \tnumQuery_{\type_t,t+1}^2}\\
        & \int_{1-\tq_{\type_t,t}^*}^{1-q_{\type_t,t}^{\pi}} \cdfInverse_{\type_t}(u)\, \dd u \leq \cdfInverse_{\type_t}(1-\tq_{\type_t,t}^*) \cdot (\tq_{\type_t,t}^*-1)+ \frac{(1-\tq_{\type_t,t}^*)^2}{2\alpha}
    \end{aligned}
\end{equation*}

Therefore, with probability at least $1-\frac{1}{\numPeriod}$, we get
\begin{equation} \label{eq:myopic_regret_case1}
    \begin{aligned}
        \myopic_t(\pi,\brconstraint) & \leq \expect[\instance_{t+1} \sim G]{\frac{(1-\tq_{\type_t,t}^*)^2}{2\alpha} + \frac{(1-\tq_{\type_t,t}^*)^2}{2\alpha \tnumQuery_{\type_t,t+1}}}\\
        & \leq \expect[\instance_{t+1} \sim G]{\frac{2(1-\hq_{\type_t,t})^2}{2\alpha} + \frac{2(\hq_{\type_t,t}-\tq_{\type_t,t}^*)^2}{2\alpha} + \frac{1}{2\alpha \tnumQuery_{\type_t,t+1}}}\\
        & \leq \frac{5\kappa^2 (\log\numPeriod)^2}{\alpha} \left(\frac{1}{s}+\frac{1}{\numPeriod-s+1}+\frac{2}{\sqrt{s(\numPeriod-s+1)}}\right) + \frac{1}{2\alpha \cdot \gamma (s-1)} 
    \end{aligned}
\end{equation}
where the second inequality follows from Basic Inequality and third inequality holds for $\hq_{\type_t,t} \geq 1-2\kappa (\frac{\log \numPeriod}{\sqrt{s}}+\frac{\log \numPeriod}{\sqrt{\numPeriod-s+1}})$, \Cref{le:q_gap} and \Cref{assump:MAP2}.

\textbf{Case 2:} when $\hq_{\type_t,t} \leq 2\kappa (\frac{\log \numPeriod}{\sqrt{s}}+\frac{\log \numPeriod}{\sqrt{\numPeriod-s+1}})$.
From \Cref{le:q_gap}, we have 
\begin{equation*}
    |\tq_{\type_t,t}^*-\hq_{\type_t,t}| \leq \kappa (\frac{\log \numPeriod}{\sqrt{s}}+\frac{\log \numPeriod}{\sqrt{\numPeriod-s+1}})
\end{equation*}
It implies that $\tq_{\type_t,t}^* \leq 3\kappa (\frac{\log \numPeriod}{\sqrt{s}}+\frac{\log \numPeriod}{\sqrt{\numPeriod-s+1}}) \leq \frac{1}{2}$ when $144\kappa^2 (\log\numPeriod)^2 \leq s \leq \numPeriod + 1 - 144\kappa^2 (\log\numPeriod)^2$. We know that query $t$ of type $\type_t$ should be rejected by our algorithm with high probability. For simplicity, we set $q_{\type_t,t}^{\pi} =0$. Thus we only need to construct a feasible solution to $\bVsemi_{\brconstraint}(\instance_{t+1})$ as it contributes negatively to our myopic regret.

We construct the following solution $\{\tq_{\type,t}''\}_{\type=1}^{\numSize}$ satisfying
\begin{equation} \label{eq:q''}
    \tq_{\type,t}'' = \tq_{\type,t}^*, \forall \type \neq \type_t \quad \text{and} \quad \tq_{\type_t,t}'' = \tq_{\type_t,t}^* \cdot \frac{\tnumQuery_{\type_t,t+1}+1}{\tnumQuery_{\type_t,t+1}}
\end{equation}

Follow the feasibility of $\{\tq_{\type,t}^*\}_{\type=1}^{\numSize}$ in (\ref{eq:qt_constraint}), we have $\sum_{\type=1}^{\numSize} \tnumQuery_{\type,t+1} \cdot a_{j,i} \cdot \tq_{\type,t}'' \leq \rconstraint_i$ for every $i \in [\numResource]$. Therefore $\{\tq_{\type,t}''\}_{\type=1}^{\numSize}$ is a feasible solution to $\bVsemi_{\brconstraint}(\instance_{t+1})$, where $\tq_{\type_t,t}'' \leq 1$ follows from $\tq_{\type_t,t}^* \leq \frac{1}{2}$ and $\tnumQuery_{\type_t,t+1} \geq 1$.

Following (\ref{eq:myopic_regret_1}), setting $q_{\type_t,t}^{\pi}=0$, we have an upper bound for myopic regret:
\begin{equation*} 
    \begin{aligned}
        & \myopic_t(\pi,\brconstraint) \\
        \leq & \expect[\instance_{t+1} \sim G]{\sum_{j=1}^\numSize \tnumQuery_{j,t+1} \cdot \int_{1-\tq_{\type,t}^*}^1 \cdfInverse_{\type}(u)\, \dd u + \int_{1-\tq_{\type_t,t}^*}^{1-q_{\type_t,t}^{\pi}} \cdfInverse_{\type_t}(u)\, \dd u  - \sum_{j=1}^\numSize \tnumQuery_{j,t+1} \cdot \int_{1-\tq_{\type,t}''}^1 \cdfInverse_{\type}(u)\, \dd u }\\
        = & \expect[\instance_{t+1} \sim G]{\tnumQuery_{\type_t,t+1} \cdot \int_{1-\tq_{\type_t,t}^*}^{1-\tq_{\type_t,t}''} \cdfInverse_{\type_t}(u)\, \dd u + \int_{1-\tq_{\type_t,t}^*}^{1-q_{\type_t,t}^{\pi}} \cdfInverse_{\type_t}(u)\, \dd u}
    \end{aligned}
\end{equation*}

Applying \Cref{le:bound_intF}, we have
\begin{equation*}
    \begin{aligned}
        & \int_{1-\tq_{\type_t,t}^*}^{1-\tq_{\type_t,t}''} \cdfInverse_{\type_t}(u)\, \dd u \leq \cdfInverse_{\type_t}(1-\tq_{\type_t,t}^*) \cdot \frac{-\tq_{\type_t,t}^*}{\tnumQuery_{\type_t,t+1}} + \frac{(\tq_{\type_t,t}^*)^2}{2\alpha \tnumQuery_{\type_t,t+1}^2}\\
        & \int_{1-\tq_{\type_t,t}^*}^{1-q_{\type_t,t}^{\pi}} \cdfInverse_{\type_t}(u)\, \dd u \leq \cdfInverse_{\type_t}(1-\tq_{\type_t,t}^*) \cdot \tq_{\type_t,t}^*+ \frac{(\tq_{\type_t,t}^*)^2}{2\alpha}
    \end{aligned}
\end{equation*}

Therefore, with probability at least $1-\frac{1}{\numPeriod}$, we get
\begin{equation} \label{eq:myopic_regret_case2}
    \begin{aligned}
        \myopic_t(\pi,\brconstraint) & \leq \expect[\instance_{t+1} \sim G]{\frac{(\tq_{\type_t,t}^*)^2}{2\alpha} + \frac{(\tq_{\type_t,t}^*)^2}{2\alpha \tnumQuery_{\type_t,t+1}}}\\
        & \leq \expect[\instance_{t+1} \sim G]{\frac{2(\hq_{\type_t,t})^2}{2\alpha} + \frac{2(\hq_{\type_t,t}-\tq_{\type_t,t}^*)^2}{2\alpha} + \frac{1}{2\alpha \tnumQuery_{\type_t,t+1}}}\\
        & \leq \frac{5\kappa^2 (\log\numPeriod)^2}{\alpha} \left(\frac{1}{s}+\frac{1}{\numPeriod-s+1}+\frac{2}{\sqrt{s(\numPeriod-s+1)}}\right) + \frac{1}{2\alpha \cdot \gamma (s-1)} 
    \end{aligned}
\end{equation}
where the second inequality follows from Basic Inequality and third inequality holds for $\hq_{\type_t,t} \leq 2\kappa (\frac{\log \numPeriod}{\sqrt{s}}+\frac{\log \numPeriod}{\sqrt{\numPeriod-s+1}})$, \Cref{le:q_gap} and \Cref{assump:MAP2}.

\textbf{Case 3:} when $2\kappa (\frac{\log \numPeriod}{\sqrt{s}}+\frac{\log \numPeriod}{\sqrt{\numPeriod-s+1}}) \leq \hq_{\type_t,t} \leq 1-2\kappa (\frac{\log \numPeriod}{\sqrt{s}}+\frac{\log \numPeriod}{\sqrt{\numPeriod-s+1}})$.
We know that
\begin{equation*}
    \brconstraint \geq \tsnumQuery_{\type_t,t} \cdot \hq_{\type_t,t} \cdot \size_{\type_t} \geq \gamma \cdot s \cdot 2\kappa \frac{\log \numPeriod}{\sqrt{s}} \cdot \size_{\type_t} \geq \size_{\type_t}
\end{equation*}
for large $s \geq \frac{1}{4 \gamma^2 \kappa^2 (\log\numPeriod)^2}$. Therefore, we always have enough remaining capacity to serve quert $t$ with type $\type_t$.
From \Cref{le:q_gap}, we have 
\begin{equation*}
    |\tq_{\type_t,t}^*-\hq_{\type_t,t}| \leq \kappa (\frac{\log \numPeriod}{\sqrt{s}}+\frac{\log \numPeriod}{\sqrt{\numPeriod-s+1}})
\end{equation*}
which implies that $\kappa (\frac{\log \numPeriod}{\sqrt{s}}+\frac{\log \numPeriod}{\sqrt{\numPeriod-s+1}}) \leq \tq_{\type_t,t}^* \leq 1-\kappa (\frac{\log \numPeriod}{\sqrt{s}}+\frac{\log \numPeriod}{\sqrt{\numPeriod-s+1}})$.
In this case, the offline optimum accept query $t$ with a probability neither close to 0 nor close to 1. Thus we set $q_{\type_t,t}^{\pi} = \hq_{\type_t,t}$ in the algorithm.

We construct the solution $\{\tq_{\type,t}'\}_{\type=1}^{\numSize}$ to $\bVsemi_{\brconstraint - \size_{\type_t}}(\instance_{t+1})$ as (\ref{eq:q'}) and the solution $\{\tq_{\type,t}''\}_{\type=1}^{\numSize}$ to $\bVsemi_{\brconstraint}(\instance_{t+1})$ as (\ref{eq:q''}). Then for $s \geq \frac{4}{\gamma^2 \kappa^2 (\log\numPeriod)^2}$, we have
\begin{equation*}
    \begin{aligned}
        & \tq_{\type_t,t}^* \geq \kappa \frac{\log \numPeriod}{\sqrt{s}} \geq \frac{1}{\gamma(s-1)} \geq \frac{1}{\tnumQuery_{\type_t,t+1}}\\
        & \tq_{\type_t,t}^* \leq 1 - \kappa \frac{\log \numPeriod}{\sqrt{s}} \leq 1 - \frac{1}{\gamma(s-1)} \leq 1 - \frac{1}{\tnumQuery_{\type_t,t+1}}
    \end{aligned}
\end{equation*}

Therefore $\tq_{\type_t,t}' \geq 0$ and $\tq_{\type_t,t}'' \leq 1$. Following the feasibility of $\{\tq_{\type,t}^*\}_{\type=1}^{\numSize}$ in (\ref{eq:qt_constraint}), we know that $\{\tq_{\type,t}'\}_{\type=1}^{\numSize}$ is feasible to $\bVsemi_{\brconstraint - \size_{\type_t}}(\instance_{t+1})$ and $\{\tq_{\type,t}''\}_{\type=1}^{\numSize}$ is feasible to $\bVsemi_{\brconstraint}(\instance_{t+1})$.

Following (\ref{eq:myopic_regret_1}) and \Cref{le:bound_intF}, by setting $q_{\type_t,t}^{\pi} = \hq_{\type_t,t}$, with probability at least $1-\frac{1}{\numPeriod}$, we have
\begin{equation} \label{eq:myopic_regret_case3}
    \begin{aligned}
        & \myopic_t(\pi,\brconstraint) \\
        \leq & \sexpect[\instance_{t+1} \sim G]{q_{\type_t,t}^{\pi} \tnumQuery_{\type_t,t+1} \int_{1-\tq_{\type_t,t}^*}^{1-\tq_{\type_t,t}'} \cdfInverse_{\type_t}(u)\, \dd u + \int_{1-\tq_{\type_t,t}^*}^{1-q_{\type_t,t}^{\pi}} \cdfInverse_{\type_t}(u)\, \dd u\\
        & + (1-q_{\type_t,t}^{\pi}) \tnumQuery_{\type_t,t+1} \int_{1-\tq_{\type_t,t}^*}^{1-\tq_{\type_t,t}''} \cdfInverse_{\type_t}(u)\, \dd u}\\
        \leq & \sexpect[\instance_{t+1} \sim G]{q_{\type_t,t}^{\pi} \frac{(1-\tq_{\type_t,t}^*)^2}{2\alpha \tnumQuery_{\type_t,t+1}} + (1-q_{\type_t,t}^{\pi}) \frac{(\tq_{\type_t,t}^*)^2}{2\alpha \tnumQuery_{\type_t,t+1}} + \frac{(\tq_{\type_t,t}^* - \hq_{\type_t,t})^2}{2\alpha}}\\
        \leq & \sexpect[\instance_{t+1} \sim G]{\frac{1}{2\alpha \tnumQuery_{\type_t,t+1}} + \frac{(\tq_{\type_t,t}^* - \hq_{\type_t,t})^2}{2\alpha}}\\
        \leq & \frac{1}{2\alpha \cdot \gamma (s-1)} + \frac{\kappa^2 (\log\numPeriod)^2}{2\alpha}(\frac{1}{s}+\frac{1}{\numPeriod-s+1}+\frac{2}{\sqrt{s(\numPeriod-s+1)}})
    \end{aligned}
\end{equation}

We define $s_0 = \max\{144\kappa^2 (\log\numPeriod)^2, \frac{4}{\gamma^2 \kappa^2 (\log\numPeriod)^2}, \frac{2}{\gamma}\}$. To conclude, when $s_0 \leq s \leq \numPeriod - s_0 +1$, combining (\ref{eq:myopic_regret_case1}), (\ref{eq:myopic_regret_case2}) and (\ref{eq:myopic_regret_case3}), with probability at least $1-\frac{1}{\numPeriod}$, we have
\begin{equation*}
    \myopic_t(\pi,\brconstraint) \leq \frac{1}{2\alpha \cdot \gamma (s-1)} + \frac{5 \kappa^2 (\log \numPeriod)^2}{\alpha}\left(\frac{1}{s}+\frac{1}{\numPeriod-s+1}+\frac{2}{\sqrt{s(\numPeriod-s+1)}}\right)
\end{equation*}

Following (\ref{eq:myopic_decompose}), we obtain the upper bound for total regret:
\begin{equation*}
    \begin{aligned}
        \regret(\pi) & \leq  \sum_{t=1}^{\numPeriod} \expect[\brconstraint_t^{\pi}]{\myopic_t(\pi,\brconstraint_t^{\pi})} + \numResource \cdot \uppreward\\
        & \leq \left(\frac{1}{\alpha \gamma} + \frac{5 \kappa^2 (\log\numPeriod)^2}{\alpha}\right)(2\log\numPeriod+2+2\pi) + 2 s_0 \cdot \uppreward + \uppreward \numPeriod \cdot \frac{1}{\numPeriod} + \numResource \cdot \uppreward\\
        & = O(\sqrt{n}\cdot (\log\numPeriod)^3 + m)
    \end{aligned}
\end{equation*}

\xhdr{Proof of \Cref{le:bound_intF}:}
Following (\ref{eq:Gj_prop}), we know that for any $\type \in [\numSize]$, $G_j''(q) \in [-\frac{1}{\alpha},-\frac{1}{\beta}]$. From the strong concavity of $G_j$, we have
\begin{equation*}
    \begin{aligned}
        \int_{1-q_1}^{1-q_2} \cdfInverse_j(u)\, \dd u & = G_j(q_1) - G_j(q_2)\\
        & \leq G_j'(q_1) \cdot (q_1-q_2) + \frac{(q_1-q_2)^2}{2\alpha}\\
        & = \cdfInverse_j(1-q_1)\cdot (q_1-q_2) + \frac{(q_1-q_2)^2}{2\alpha}
    \end{aligned}
\end{equation*}

\end{APPENDICES}

\end{document}